\documentclass{article}
\UseRawInputEncoding
\usepackage{microtype}
\usepackage{graphicx}
\usepackage{subcaption}
\usepackage{booktabs} 
\usepackage{threeparttable}
\usepackage{siunitx}
\usepackage[dvipsnames]{xcolor}
\usepackage{multirow}
\usepackage{xurl}
\usepackage{listings}
\usepackage{xcolor}
\usepackage[most]{tcolorbox}
\usepackage{fancyvrb}   
\usepackage{rotating}
\usepackage{colortbl}
\definecolor{codegray}{rgb}{0.45,0.45,0.45}
\definecolor{codegreen}{rgb}{0.0,0.45,0.25}
\definecolor{promptblue}{rgb}{0.12,0.35,0.75}
\definecolor{lightgray}{gray}{0.9}
\usepackage{hyperref}
\lstdefinestyle{frame}{
  basicstyle=\ttfamily\footnotesize,
  keywordstyle=\color{codegreen}\bfseries,
  commentstyle=\color{codegray}\itshape,
  stringstyle=\color{codegreen},
  numbers=none,
  numberstyle=\tiny\color{codegray},
  stepnumber=1,
  numbersep=6pt,
  frame=single,
  framesep=5pt,
  breaklines=true,
  showstringspaces=false,
  tabsize=2
}

\lstdefinestyle{clean}{
  basicstyle=\ttfamily\small,
  keywordstyle=\color{codegreen}\bfseries,
  commentstyle=\color{codegray}\itshape,
  stringstyle=\color{codegreen},
  numbers=none,          
  frame=none,            
  breaklines=true,       
  showstringspaces=false,
  tabsize=2,
  columns=fullflexible
}
\newtcolorbox{topblueprompt}[2][]{
  enhanced,
  breakable,
  colback=white,
  colframe=promptblue,
  boxrule=0pt,
  toprule=1.5pt,
  bottomrule=0pt,
  leftrule=0pt,
  rightrule=0pt,
  left=6pt,
  right=6pt,
  top=6pt,
  bottom=6pt,
  title={#2},
  coltitle=white,
  fonttitle=\bfseries,
  sharp corners,
  listing only,
  listing options={style=clean},
  #1
}

\usepackage{fancyvrb}
\usepackage[preprint]{icml2026}

\usepackage{amsmath}
\usepackage{amssymb}
\usepackage{mathtools}
\usepackage{amsthm}
\usepackage[normalem]{ulem}

\usepackage[capitalize,noabbrev]{cleveref}

\theoremstyle{plain}

\theoremstyle{definition}

\theoremstyle{remark}

\newcommand{\gainp}[1]{\scriptsize\textcolor{ForestGreen}{+#1}}
\newcommand{\gainn}[1]{\scriptsize\textcolor{BrickRed}{#1}}
\usepackage{xspace}
\newcommand{\ouralgo}{ALMA\xspace} 
\usepackage[font=small,labelfont=bf]{caption}
\newif\ifcomment
\commentfalse

\usepackage[textsize=tiny]{todonotes}

\icmltitlerunning{Learning to Continually Learn via Meta-learning Agentic Memory Designs}

\begin{document}

\twocolumn[
  \icmltitle{Learning to Continually Learn via Meta-learning Agentic Memory Designs}



  \icmlsetsymbol{equal}{*}

  \begin{icmlauthorlist}
    \icmlauthor{Yiming Xiong}{ubc}
    \icmlauthor{Shengran Hu}{ubc,vector}
    \icmlauthor{Jeff Clune}{ubc,vector,cifar}
  \end{icmlauthorlist}

  \icmlaffiliation{ubc}{University of British Columbia}
  \icmlaffiliation{vector}{Vector Institute}
  \icmlaffiliation{cifar}{Canada CIFAR AI Chair}

  \icmlcorrespondingauthor{Yiming Xiong}{yiming33@student.ubc.ca}

  \icmlkeywords{Machine Learning, ICML, Meta-Learning, Agentic System, Continual Learning, Agentic Memory}

  \vskip 0.3in
]



\printAffiliationsAndNotice{}  

\begin{abstract}
The statelessness of foundation models bottlenecks agentic systems’ ability to continually learn, a core capability for long-horizon reasoning and adaptation. To address this limitation, agentic systems commonly incorporate memory modules to retain and reuse past experience, aiming for continual learning during test time. However, most existing memory designs are human-crafted and fixed, which limits their ability to adapt to the diversity and non-stationarity of real-world tasks. In this paper, we introduce \ouralgo (\textbf{A}utomated meta-\textbf{L}earning of \textbf{M}emory designs for \textbf{A}gentic systems), a framework that meta-learns memory designs to replace hand-engineered memory designs, therefore minimizing human effort and enabling agentic systems to be continual learners across diverse domains. Our approach employs a Meta Agent that searches over memory designs expressed as executable code in an open-ended manner, theoretically allowing the discovery of arbitrary memory designs, including database schemas as well as their retrieval and update mechanisms. Extensive experiments across four sequential decision-making domains demonstrate that the learned memory designs enable more effective and efficient learning from experience than state-of-the-art human-crafted memory designs on all benchmarks. When developed and deployed safely, \ouralgo represents a step toward self-improving AI systems that learn to be adaptive, continual learners. 
All code is open-sourced at \url{https://github.com/zksha/alma.git}.
\end{abstract}

\section{Introduction}

Agentic systems powered by Foundation Models (FMs) have enabled autonomous decision-making and task execution across a wide range of domains \citep{yao2023react, wang2023voyager, John2024SWE}. However, the stateless nature of FMs during inference challenges agentic systems to accumulate experience and continually learn from past interactions, causing them to repeatedly solve tasks from scratch and limiting their ability to improve over time \citep{child2019generatinglongsequencessparse,Aydar2022Recurrent}. Memory addresses this limitation by enabling agentic systems to continually store and reuse past experiences \citep{Zhong2024MemoryBank, packer2023memgpt}. This allows accumulated experience to inform decision-making in future tasks, aiming for continual learning over time.

However, memory design, the architectural specification that determines how memories are represented, stored, retrieved, and updated, is still predominantly handcrafted by humans. Diverse domains require distinct memory designs to leverage unique aspects of experience, resulting in human researchers manually tailoring a wide array of designs to specific tasks \citep{zhang2024survey,hu2025memory}. 
For example, in conversational agents, memory may focus on retaining facts about the user, such as preferences and personal information~\citep{chhikara2025mem0,rasmussen2025zep}. In contrast, for strategic games, memory should extract abstract skills and strategies from previous interactions rather than details that may change across episodes~\citep{tang2025agent, ouyang2025reasoningbank}.
Therefore, manually identifying an optimal memory design for each domain is both difficult and labor-intensive.

A recurring theme in machine learning history is that handcrafted components in AI systems are eventually replaced by learned, more effective ones \citep{clune2019ai, hutter2019automated, sutton2019bitter}. A classic example of this paradigm is the transition from hand-designed features to learned representations in computer vision \citep{Alex2017ImageNet}. More recently, the trend of ``learning to learn'' has expanded to learning neural architectures \citep{zoph2017neural}, optimization algorithms \citep{Marcin2016Learning}, training environments \citep{Rui2019POET, Zhang2024OMNI, Faldor2025OMNI}, and designs for agentic systems \citep{hu2024ADAS,zhang2025darwin}, among others. In this paper, we investigate whether agentic systems can learn to continually learn by automating the design of their memory components. This enables memory designs to be specialized for diverse domains without relying on manual engineering.

\begin{figure*}[t]
  \begin{center}
    \centerline{\includegraphics[width=0.83\textwidth]{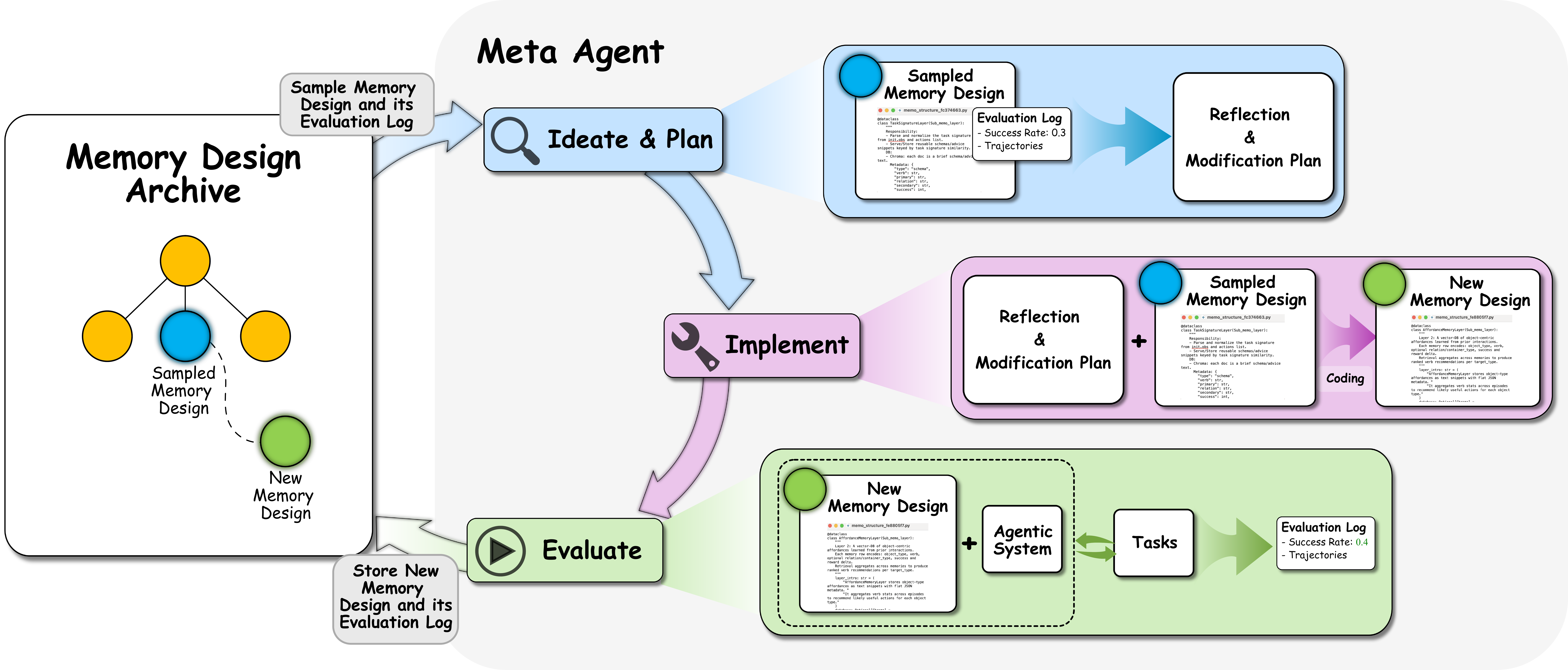}}
        \captionsetup{
      aboveskip=3pt,  
      belowskip=-2pt  
    }
    \caption{
\textbf{Open-ended Exploration Process of \ouralgo.}
The Meta Agent first ideates and proposes a plan by reflecting on the code and evaluation logs of the sampled memory design. It then implements the plan by programming the new design in code. Finally, it verifies the correctness of the new memory design and evaluates it with an agentic system. The evaluated memory design is subsequently added to the memory design archive for future sampling.
    }
    \label{main_workflow}
  \end{center}
\end{figure*}

We propose \ouralgo (\textbf{A}utomated meta-\textbf{L}earning of \textbf{M}emory designs for \textbf{A}gentic systems), a framework that adopts a Meta Agent to explore novel memory designs for agentic systems through open-ended exploration. 
We use code as the search space, giving \ouralgo the theoretical potential to build any memory design.
As shown in \cref{main_workflow}, the Meta Agent samples previously explored memory designs from an archive, where all explored memory designs and their evaluation logs are stored. 
Then the Meta Agent reflects on the sampled memory designs to generate new ideas and plans, which are implemented in code.
The new designs are subsequently validated and evaluated, with the resulting logs added back to the archive to guide future sampling.

We evaluate our method across four sequential decision-making domains, including ALFWorld \citep{shridhar2021Alfworld}, TextWorld \citep{textworld}, Baba Is AI \citep{cloos2024baba}, and MiniHack \citep{samvelyan2021minihack}. These domains serve as testbeds for evaluating an agent's capability to continually learn from experience. In these domains, agents benefit from storing and reusing their experiences to acquire effective strategies and useful information that is absent from the FMs' pre-trained knowledge. The results show that \ouralgo can discover memory designs that are tailored for the needs of different domains, consistently surpassing state-of-the-art human-designed memory baselines (\Cref{tab:testing_results}). Beyond superior performance, our learned memory designs are more cost-efficient than most human-designed memory baselines (\Cref{tradeoff_combined}). Furthermore, the learned memory designs scale performance more effectively with memory size and learn faster when facing task distribution shifts (\Cref{scale_plot,trans_plot}). Overall, the memory designs learned by \ouralgo demonstrate a superior ability to help agentic systems continually learn, paving the way towards developing lifelong learning agents in dynamic environments.
\vspace{-0.6em}
\section{Related Work}

\paragraph{Memory for Agentic Systems.} Memory enables agentic systems to continually learn despite the statelessness of FMs \citep{hu2025memory, Zhang2025Survey}. Existing work has categorized memory mechanisms into token-level memory, parametric memory, and latent memory \citep{hu2025memory}. Token-level memory defines databases and update rules to store information and extract insights from interaction trajectories between agents and environments. When addressing a new task, the system retrieves relevant experiences via defined procedures, which are then appended to the prompts of agentic systems as context \citep{chhikara2025mem0, rasmussen2025zep, nan2025nemori, zhang2025g, ouyang2025reasoningbank, suzgun2025DynamicCheatsheet, zhaoExpeL2024}. For example, G-Memory \citep{zhang2025g} extracts experiences and stores them in a graph-based database, and retrieves relevant experiences for new tasks via graph traversal. Reasoning Bank \citep{ouyang2025reasoningbank} performs task-wise experience generation during memory updates, while Dynamic Cheatsheet \citep{suzgun2025DynamicCheatsheet} includes accumulation of insights across trajectories. Current works also explore training models to select from predefined operations for storing and reusing experience in memory \citep{zhou2025memento, yan2025memory, liang2026learning, wang2025mem}. Other categories of memory include parametric and latent memory, which encode experience implicitly within the model's weights \citep{Zhen2024Speed, zhang2025agentlearningearlyexperience} or latent hidden states \citep{Yu2024MEMORYLLM, zhang2025memgen, zou2025latent} rather than relying on explicit text-based retrieval. The ``Learning to learn'' paradigm can theoretically support the learning of all categories of memory. In this paper, we focus on token-level memory, as it allows for faster evaluation without additional model training and provides more interpretable, transferable memory representations.

\paragraph{AI-generating algorithms and the ``Learning to learn'' paradigm.}
AI-generating algorithms \citep{clune2019ai} and automated machine learning \citep{hutter2019automated} aim to replace hand-engineered components with automatically learned ones. The approach has three key pillars: (1) meta-learning architectures, (2) meta-learning learning algorithms, and (3) the automatic generation of learning environments \citep{clune2019ai}.
Neural Architecture Search \citep{Thomas2019Neural, Lu2019NSGA, hu2023accelerating} primarily contributes to the first pillar by automating the discovery of neural architectures. Research in MAML \citep{Finn2017Agnostic}, Meta-RL \citep{wang2016learning, duan2016rl, Ben2024First}, and recent Automated Algorithm Design \citep{Liu2024EoH,liu2025systematic} has explored the second pillar by focusing on ``Learning to learn'', developing algorithms that enable agents to adapt their learning behavior across tasks \citep{Beaulieu2020Learning,Lu2024Discovering}. 
The third pillar includes approaches such as POET \citep{Rui2019POET, Rui2024EnhancedPOET}, as well as more recent work on dynamic environment generation with agentic systems \citep{Fuma2024LLM, guo2025genenv,Zhang2024OMNI}.
Our work is more related to the second pillar, as \ouralgo learns how memory designs can better enable agentic systems to continually learn. Furthermore, recent AI-generating algorithms and automated machine learning approaches have leveraged FMs to explore new components in AI systems through code generation \citep{hu2024ADAS, zhang2025darwin, Faldor2025OMNI, yamada2025ai,lu2024ai, Lu2024Discovering,lu2025automated, Lehman2024}. 
Similarly, we adopt a Meta Agent powered by FMs to propose novel memory designs by programming in code.


\paragraph{Automated Design of Agentic Systems.}
Previous work has studied the automated design of agentic systems to learn better components in agentic systems \citep{ hu2024ADAS,Zhang2025aflow,zhou2024symbolic,yin2025godel,zhuge24a, rosser2025agentbreeder, zhang2025multiagent, Yu2025AgentSquare, zhang2025swarmagentic, zhang2025multiagent, ye2025masgpt, zhang2025darwin}. Since recent works such as ADAS \citep{hu2024ADAS} employ code as the search space for agentic system designs, they theoretically enable the learning of the memory component. For example, AgentSquare \citep{Yu2025AgentSquare} attempts to co-learn memory designs along with other components like tools in agentic systems.  
However, while existing works evaluate only the one-shot performance of agentic systems, \ouralgo explicitly optimizes for a memory design's capability to facilitate continual learning from past experience, enabling the discovery of more effective memory designs.
A concurrent work to \ouralgo also explores an approach for learning memory designs \citep{zhang2025memevolve}. However, it relies on the initialization with many existing handcrafted memory designs and greedy selection of top-performing designs, which limits open-ended exploration. 
In contrast, \ouralgo adopts an open-ended exploration of memory designs and learns from scratch. Open-ended exploration is shown to be important to search for high-performance structure \citep{Joel2008Exploiting,Lehman2011Evolutionary,Lehman2011Evolving,Conti2018Improving, Stanley2019Why}. Therefore, incorporating open-ended exploration into the search for optimal memory designs presents a promising avenue. Our ablation studies show that indeed open-ended exploration can learn better memory designs than greedy-selection-based optimization (Appendix~\ref{app:greedy_explore}).

\vspace{-0.4em}
\section{Learning of Memory Designs}

We propose \ouralgo, a framework for learning memory designs within a code-based search space via open-ended exploration, discovering memory designs that effectively collect, summarize, and reuse experience.
Specifically, a Meta Agent \citep{hu2024ADAS} iteratively proposes a new candidate memory design by reflecting on the results of previously discovered designs, implements the new design through code generation and debugging, and evaluates its performance by integrating it into an agentic system. 
The pseudocode of our learning process is provided in Appendix~\ref{Pseudocode}.



\vspace{-0.5em}
\subsection{Search Space for Memory Designs} 
The search space defines all memory designs that can be represented and discovered. In this paper, we use code as the search space, thus theoretically allowing all possible memory designs to be discovered due to the Turing completeness of many programming languages, i.e., Python in our case.
Representing memory designs in code also enables interpretability and allows FMs to leverage their prior knowledge acquired during pretraining about agentic systems and coding \citep{hu2024ADAS}.
Due to the vastness of the code-based search space, it is inefficient for the Meta Agent to construct all basic functions of a memory design from scratch. 
We therefore provide the Meta Agent a simple abstraction that facilitates exploration while preserving flexibility. 
Our abstraction design is inspired by common patterns observed in existing handcrafted memory designs \citep{nan2025nemori, chhikara2025mem0, ouyang2025reasoningbank, zhang2025g, tang2025agent, suzgun2025DynamicCheatsheet}. The abstraction allows the memory module to interact with the agentic system through two primary interfaces: \texttt{general\_update()} and \texttt{general\_retrieve()}. After collecting new interactions with the environment, the agent calls \texttt{general\_update()} to extract useful experience into memory. When faced with a new task, the agent calls \texttt{general\_retrieve()} to access relevant experience. 
Internally, each interface can coordinate multiple sub-modules. 
Each sub-module implements its \texttt{update()} and \texttt{retrieve()} logic, 
and optionally maintains its own database when needed. The output of one sub-module can serve as input for subsequent sub-modules, enabling hierarchical and modular memory designs.
For example, ReasoningBank \citep{ouyang2025reasoningbank} employs a vector-database sub-module to store task descriptions, along with an additional sub-module that extracts experience learned from individual tasks with FMs. 
Similarly, G-memory \citep{zhang2025g} incorporates a graph-database sub-module to model linkages between extracted insights and a vector-database sub-module to store task descriptions. 
We implement the described abstraction in Python Abstract Classes, and the code is available in Appendix~\ref{search_space}.
\vspace{-0.5em}
\subsection{Evaluation of Memory Designs} 
\label{memory eval}
Based on existing work \citep{nan2025nemori, chhikara2025mem0, ouyang2025reasoningbank, zhang2025g, tang2025agent, suzgun2025DynamicCheatsheet, hu2025memory, wang2024agent, Zhong2024MemoryBank, packer2023memgpt, fang2025memp, zhaoExpeL2024}, we summarize that memory components in agentic systems generally operate in two phases.

\noindent\textbf{Memory Collection Phase.}\hspace{0.5em}
The goal of this phase is not task success, but rather collecting knowledge and updating memory for later use in the Deployment Phase. There are two common ways to obtain raw trajectories for updating memory at this stage: (1) leveraging existing agent trajectories from available datasets \citep{Zhong2024MemoryBank, chhikara2025mem0}, or (2) starting from an empty memory state and collecting trajectories by running the agent on tasks \citep{fang2025memp, tang2025agent}. This phase typically does not involve retrieving from memory \citep{Zhong2024MemoryBank, chhikara2025mem0, wang2024agent, packer2023memgpt, rasmussen2025zep}.

\noindent\textbf{Deployment Phase.}\hspace{0.5em}
This is the main stage where the agentic system is deployed to applications and expected to achieve task success. Given the memory collected after the Memory Collection Phase, the agentic system retrieves from memory for each incoming task and attempts to solve it. This stage can operate in two modes: \textbf{dynamic}, which updates memory with newly collected trajectories from incoming tasks~\citep{zhang2025g, ouyang2025reasoningbank}, or \textbf{static}, which keeps memory fixed~\citep{tang2025agent, rasmussen2025zep}.

We evaluate each memory design by first running the Memory Collection Phase and then recording the success rate during the Deployment Phase, using an identical, fixed agentic system, with the memory design being evaluated. The two modes in the Deployment Phase serve different purposes: static mode assesses how effectively the agent leverages a fixed memory to solve new tasks, while dynamic mode measures how well a memory design adapts to a new task distribution through dynamic updates and retrieval. When learning memory designs, we use only static mode to reduce variance during evaluations. The \texttt{general\_update} and \texttt{general\_retrieve} modules contribute to the performance of the memory design through memory collection and knowledge retrieval, respectively. When testing the best-learned memory design, we evaluate under both modes.
More details on evaluations are available in Appendix~\ref{eval_process}.
\vspace{-0.5em}
\subsection{Open-Ended Exploration} 
Similar to ADAS \citep{hu2024ADAS}, we propose a framework for open-ended exploration of memory designs with a Meta Agent programming novel memory designs in code. 
The algorithm proceeds as follows: (1) An archive is initialized with Python abstract classes as an empty memory design template. (2) The Meta Agent samples previously discovered designs from the archive, reflects on their performance outcomes, and proposes new memory designs. (3) The newly proposed designs are evaluated on the target domain. If errors occur during evaluation, the Meta Agent performs self-reflection to refine the design, repeating this process up to three times if necessary. (4) Finally, the memory design is added to the archive along with its evaluation results and logs, and the process continues with the updated archive until the maximum number of iterations is reached.

\noindent\textbf{Sampling Memory Designs.}\hspace{0.5em} 
We sample memory designs from the archive to serve as stepping stones for the exploration of new memory designs at each step.
Similar to DGM~\citep{zhang2025darwin}, each design in the archive is assigned a sampling probability roughly proportional to its success rate and inversely proportional to the number of times it is sampled. This prioritizes designs that perform well but have been sampled less frequently, enabling a balance between refining successful designs and exploring underexplored candidates. All designs maintain non-zero sampling probabilities, keeping all potential improvements reachable.
Details are provided in Appendix~\ref{selection_process}.

\noindent\textbf{Proposing New Memory Design by the Meta Agent.}\hspace{0.5em} 
After sampling previous memory designs from the archive, the Meta Agent reflects on their evaluation results and proposes new memory designs by implementing them in code.
Specifically, the Meta Agent first proposes ideas, then plans for the new designs through analyzing the sampled designs along with their success rate and interaction logs.
Based on the plans, the Meta Agent implements the new memory designs in code and performs trial runs. Runtime errors during the trial runs trigger debugging reflections to refine the implementations.
Finally, the new memory designs are evaluated. Their evaluation logs are added to the memory design archive to support future sampling. 
Prompts for planning, implementation, and debugging are in Appendix~\ref{prompts_meta}.
\vspace{-0.6em}
\section{Experiments}
\label{Experiment}

\subsection{Experiment Setup}
\label{Exp setup}
We perform \ouralgo on four sequential decision-making benchmarks (Section~\ref{Benchmarks}).
For each benchmark, the dataset is divided into a learning set and a testing set, which are used for memory design learning and for the testing of the learned memory designs to ensure testing on unseen data.  
Both sets are further split evenly, with the first half used for the Memory Collection Phase and the second half for the Deployment Phase (\Cref{memory eval}). To reduce variance, we execute the Deployment Phase three times for learning and testing, and report the average success rate with the standard error.
For clarity, in the following sections, ``success rate'' refers to the average success rate over three runs. 

At each learning step, we sample up to five memory designs from the archive without replacement. We expect the choice between sampling with or without replacement to have minimal impact (Appendix~\ref{app:learning_process}).
We run \ouralgo for 11 learning steps to discover 43 memory designs.

During the learning of memory designs, \texttt{GPT-5} is used in the Meta Agent, while \texttt{GPT-5-nano} is used in the fixed agentic system for evaluating memory designs. While the Meta Agent can choose to use any model (or train models) as FMs within memory designs, we provide \texttt{GPT-4o-mini}, \texttt{GPT-4.1}, and \texttt{text-embedding-3-small} as tools for the Meta Agent to explore workflows such as extracting insights or computing semantic similarity. More details on the learning process setup are available in Appendix~\ref{app:learning_process}.

During the testing of the learned memory designs, we test under two agentic system settings: the first employs \texttt{GPT-5-nano}, matching the learning setup. The second replaces it with a more capable FM, \texttt{GPT-5-mini}, to assess the transferability of memory designs to stronger agentic systems (\Cref{tab:testing_results}). We use \texttt{GPT-4o-mini} and \texttt{text-embedding-3-small} as FMs across all memory designs during testing to ensure a fair comparison. More details on the testing setup are available in Appendix~\ref{app:testing_details}.

\begin{table*}[ht]
\caption{\textbf{Comparison of success rates between the learned memory designs and baseline memory designs across environments.}
\textbf{Top}: Learning and testing the memory designs with \texttt{GPT-5-nano} as the FM in the agentic system. \textbf{Bottom}: Transferring the memory designs to test with \texttt{GPT-5-mini} in the agentic system. In both cases, learned memory designs outperform human-designed baselines, demonstrating that they are more effective than human-designed memory and generalize robustly across different FMs.
Results are reported as mean $\pm$ standard error in percentages, calculated over three runs of the Deployment Phase. The relative gains over the no-memory baseline are color-coded. 
}
\label{tab:testing_results}
\begin{center}
\footnotesize
\setlength{\tabcolsep}{4pt}

\begin{tabular}{l
  S[table-format=3.1] @{\,\ensuremath{\pm}\!} S[table-format=2.1]@{\hspace{0.3em}} l
  S[table-format=3.1] @{\,\ensuremath{\pm}\!} S[table-format=2.1]@{\hspace{0.3em}} l
  S[table-format=3.1] @{\,\ensuremath{\pm}\!} S[table-format=2.1]@{\hspace{0.3em}} l
  S[table-format=3.1] @{\,\ensuremath{\pm}\!} S[table-format=2.1]@{\hspace{0.3em}} l
  S[table-format=3.1]@{\hspace{0.3em}} l}

\toprule
\textbf{Memory Designs} &
\multicolumn{3}{c}{\textbf{ALFWorld}} &
\multicolumn{3}{c}{\textbf{Textworld}} &
\multicolumn{3}{c}{\textbf{Baba Is AI}} &
\multicolumn{3}{c}{\textbf{MiniHack}} &
\multicolumn{2}{c}{\textbf{Overall Avg.}} \\
\midrule

\multicolumn{15}{c}{\textbf{GPT-5-nano}} \\
\midrule

No Memory & 2.9 & 0.8 & {} & 5.4 & 1.0 & {} & 9.5 & 2.4 & {} & 6.7 & 1.7 & {} & 6.1 & {} \\

\rowcolor{lightgray}
\multicolumn{15}{c}{\textbf{Manual Memory Designs}} \\

Trajectory Retrieval & 5.2 & 1.9 & \gainp{2.3} & 2.7 & 0.9 & \gainn{-2.7} & \bfseries 19.0 & 6.3 & \gainp{9.5} & 7.5 & 1.4 & \gainp{0.8} & 8.6 & \gainp{2.5} \\
Reasoning Bank & 5.2 & 1.3 & \gainp{2.3} & 5.3 & 0.8 & \gainn{-0.1} & 9.5 & 2.4 & \gainp{0.0} & 9.8 & 1.7 & \gainp{3.7} & 7.5 & \gainp{1.4} \\
Dynamic Cheatsheet & 5.7 & 0.8 & \gainp{2.8} & 4.3 & 1.0 & \gainn{-1.1} & 9.5 & 4.8 & \gainp{0.0} & 9.2 & 0.8 & \gainp{2.5} & 7.2 & \gainp{1.1} \\
G-Memory & 7.6 & 0.5 & \gainp{4.7} & 2.1 & 0.8 & \gainn{-3.3} & 14.3 & 4.1 & \gainp{4.8} & 6.8 & 2.3 & \gainp{0.1} & 7.7 & \gainp{1.6} \\

\rowcolor{lightgray}
\multicolumn{15}{c}{\textbf{Learned Memory Designs}} \\

\ouralgo (Ours) & \bfseries 12.4 & 0.5 & \gainp{9.5} & \bfseries 6.2 & 1.7 & \gainp{0.8} & \bfseries 19.0 & 2.4 & \gainp{9.5} & \bfseries 11.7 & 2.2 & \gainp{5.0} & \bfseries 12.3 & \gainp{6.2} \\

\midrule
\midrule

\multicolumn{15}{c}{\textbf{GPT-5-mini}} \\
\midrule

No Memory & 67.6 & 1.0 & {} & 60.5 & 4.4 & {} & 21.4 & 0.0 & {} & 15.0 & 1.4 & {} & 41.1 & {} \\

\rowcolor{lightgray}
\multicolumn{15}{c}{\textbf{Manual Memory Designs}} \\

Trajectory Retrieval & 80.0 & 1.4 & \gainp{12.4} & 67.0 & 1.3 & \gainp{6.5} & 30.9 & 2.4 & \gainp{9.5} & 16.7 & 0.8 & \gainp{1.7} & 48.6 & \gainp{7.5} \\
Reasoning Bank & 67.1 & 2.9 & \gainn{-0.5} & 56.1 & 3.4 & \gainn{-4.4} & 21.4 & 4.1 & \gainp{0.0} & 15.8 & 3.0 & \gainp{0.8} & 40.1 & \gainn{-1.0} \\
Dynamic Cheatsheet & 78.6 & 3.3 & \gainp{11.0} & 57.8 & 7.9 & \gainn{-2.7} & \bfseries 38.0 & 6.3 & \gainp{16.6} & 11.7 & 3.3 & \gainn{-3.3} & 46.5 & \gainp{5.4} \\
G-Memory & 74.8 & 0.5 & \gainp{7.2} & 68.8 & 4.7 & \gainp{8.3} & 26.2 & 6.3 & \gainp{4.8} & 14.2 & 2.2 & \gainn{-0.8} & 46.0 & \gainp{4.9} \\

\rowcolor{lightgray}
\multicolumn{15}{c}{\textbf{Learned Memory Designs (Transferred)}} \\

\ouralgo (Ours) & \bfseries 87.1 & 1.4 & \gainp{19.5} & \bfseries 75.0 & 2.3 & \gainp{14.5} & 33.3 & 2.4 & \gainp{11.9} & \bfseries 20.0 & 2.9 & \gainp{5.0} & \bfseries 53.9 & \gainp{12.8} \\

\bottomrule
\end{tabular}
\end{center}
\end{table*}

\vspace{-0.5em}
\subsection{Benchmarks} 
\label{Benchmarks}
We adopt four sequential decision-making benchmarks with varying levels of difficulty \citep{balrog}.
In all benchmarks, agentic systems perceive the environment through text and interact with it by generating actions in natural language.
The benchmarks are well-suited for evaluating memory designs because they require acquiring and utilizing experience that does not exist in the pre-trained knowledge of FMs \citep{balrog}.
We test \ouralgo on four benchmarks: 
(1) ALFWorld~\citep{shridhar2021Alfworld}, a text-based simulation of embodied household tasks that challenges agents to ground language instructions to action sequences in kitchen environments; (2) TextWorld~\citep{textworld}, text adventure games that require systematic exploration and reasoning to navigate partially observable environments; (3) Baba Is AI~\citep{cloos2024baba}, a strategic puzzle game where agents must manipulate the rules of the game, requiring complex reasoning and adaptation to changing game mechanics; (4) MiniHack~\citep{samvelyan2021minihack}, a simplified version of NetHack~\citep{kuttler2020nle} that challenges agents with long-horizon decision-making in procedurally generated dungeon environments.

For ALFWorld, we are using the default setting \citep{shridhar2021Alfworld}. For all other benchmarks, we use BALROG \citep{balrog}, following its settings and prompt for interactions during learning and final testing. More details on benchmarks are available in Appendix~\ref{app:benchmark}.
\vspace{-0.5em}
\subsection{Baselines} 
\label{Baselines}
We select four representative state-of-the-art human-designed memory systems as baselines. These baselines are entirely handcrafted, with all aspects of the design (e.g., what to store, how to update, and how to retrieve) manually specified and inserted retrieved knowledge into the context of agentic systems, representing various mainstream approaches in prior work. More details are available in Appendix~\ref{app:baseline}.

\textbf{Trajectory Retrieval.} A straightforward but strong baseline that retrieves similar trajectories \citep{Park2023Generative, Xu2024International}. Task descriptions and trajectories are stored during memory update. During memory retrieval, the most relevant past trajectory is retrieved as knowledge based on the similarity of the task description embedding.
    
\textbf{ReasoningBank.} The memory designs \citep{ouyang2025reasoningbank} organize and extract experience on a per-task basis, enabling the retrieval of experience relevant to each task. During memory update, experiences are extracted from each trajectory by FMs and stored together with the corresponding task description. The experiences corresponding to the most similar task description are retrieved as knowledge.

\textbf{Dynamic Cheatsheet (Cumulative).} A global semantic memory incrementally accumulates experience by FMs across all past trajectories \citep{suzgun2025DynamicCheatsheet}. The trajectories are sequentially used to update a global cheatsheet, which works as knowledge across all tasks.
    
\textbf{G-memory.} A hierarchical graph-based memory design \citep{zhang2025g}. During the update of memory, it builds a hierarchical memory where insights and key steps in trajectories are connected in a graph based on task descriptions. When facing new tasks, it traverses the graph to retrieve relevant insights and key steps as knowledge.
\vspace{-0.5em}
\subsection{Results}
\label{Results}

\begin{figure*}[t]
  \begin{center}
    \centerline{\includegraphics[width=0.9\textwidth]{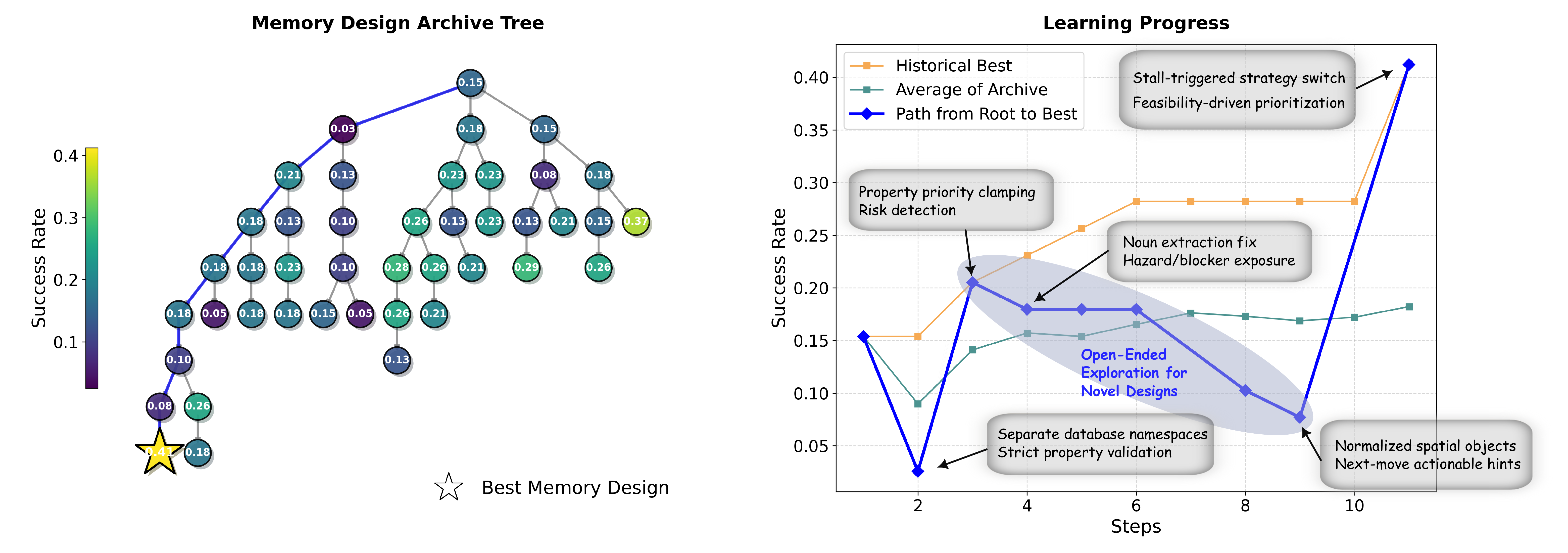}}
    \captionsetup{aboveskip=5pt, belowskip=-2pt}
    \caption{
    \textbf{The learning process of \ouralgo on Baba Is AI}, using \texttt{GPT-5-nano} as the FM in an agentic system. The learning processes of other benchmarks are shown in Appendix~\ref {app:meta_learn_results}.
    \textbf{Left:} The memory design archive tree, where each node represents a memory design produced during the open-ended exploration for ever-better memory designs. Node colors indicate the success rate, and edges indicate that each child node is derived from its parent. The memory design with the highest success rate is used as the final learned memory design.
    \textbf{Right:} The step-wise learning progress. \ouralgo progressively discovers memory designs by building on an ever-growing archive of previous discoveries. The path from the root memory design to the best memory design highlights the importance of open-ended exploration, where designs with moderate success rates serve as stepping stones toward optimal solutions. 
    }
    \label{babaisai_train}
  \end{center}
\end{figure*}

For each benchmark, we select the memory design achieving the highest success rate during the learning of memory designs as the best-learned memory design. We then evaluate both human-designed baselines and the best-learned memory designs using the testing process described in \Cref{Exp setup} with static mode in the Deployment Phase (\Cref{memory eval}).

As shown in \cref{tab:testing_results} Top, with the agentic system powered by \texttt{GPT-5-nano}, the learned memory designs achieve an overall improvement of 6.2\% compared to the no-memory baseline, and outperform all state-of-the-art human-designed memory baselines.
These improvements are consistent across benchmarks, demonstrating that learned memory designs can more effectively store and reuse past experience to continually learn than manual designs.

We further evaluate learned memory designs by changing the FM in the agentic system from \texttt{GPT-5-nano} to \texttt{GPT-5-mini} (\Cref{tab:testing_results} Bottom).
The learned memory designs improve the overall average success rate by 12.8\% over the no-memory baseline and outperform all human-designed baselines. These results demonstrate that the discovered memory designs generalize across different FMs, indicating that the improvements are robust and not tied to a specific model. The larger improvement observed with more capable FMs (12.8\% vs 6.2\%) yields a delta of 6.6\%. This delta exceeds those of all human-designed memory baselines, suggesting that our learned memory designs provide stronger support to more capable agentic systems.

To provide a closer look at the learning process, we show the learning process on Baba Is AI and the tree visualization of the resulting memory design archive (\cref{babaisai_train}).
The Meta Agent explores diverse new designs on the path toward the best-learned memory design, building upon designs with moderate success rates that nevertheless have the potential to evolve into the optimal memory design. As shown in \cref{babaisai_train}, during open-ended exploration, the Meta Agent incrementally introduces mechanisms like property validation and spatial object normalization.
While these mechanisms may not yield immediate performance gains, they serve as stepping stones that contribute to the best-learned memory design when key mechanisms (e.g., strategy switching) are introduced. 
Analyses and visualizations of the learning processes for the other benchmarks are available in the Appendix ~\ref{app:meta_learn_results}.

To evaluate the effectiveness of the open-ended exploration in \ouralgo, we conduct an ablation study comparing against greedy search, which always samples the design with the highest success rate to propose new designs. More details on greedy search are available in Appendix~\ref{app:learning_process}. We perform the learning process with greedy search on ALFWorld, resulting in success rates of 11.9\% and 77.1\% when testing the best-learned memory design with the agentic system powered by \texttt{GPT-5-nano} and \texttt{GPT-5-mini}, respectively (Appendix~\ref{app:greedy_explore}). In both cases (for \texttt{GPT-5-nano}, 11.9\% vs 12.4\%; for \texttt{GPT-5-mini}, 77.1\% vs 87.1\%), success rates are lower than those achieved by designs learned with open-ended exploration, demonstrating the benefits of an exploration-driven learning process over greedy search.

\begin{figure*}[htbp]
  \begin{center}
  \centerline{\includegraphics[width=\textwidth]{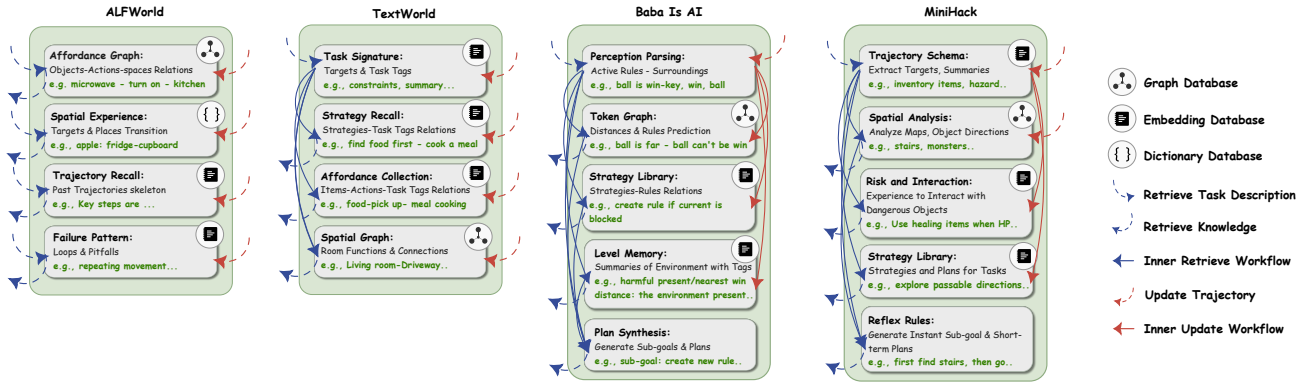}}
  \caption{
    \textbf{The visualization of the best-learned memory designs across different benchmarks.} Each sub-module in memory designs may have a dedicated database or none, depending on its function, and arrows show the retrieval and update workflows in memory designs. The name and explanation of each sub-module are generated by Meta Agent and manually summarized, respectively. Example code and output for a learned memory design are provided.
  }
  \label{meta_learn_layer}
  \end{center}
\end{figure*}

We visualize the learned memory designs for each domain (\cref{meta_learn_layer}), showing that \ouralgo discovers effective memory structures adapted to diverse task requirements. For games with explicit object interaction goals (e.g., ALFWorld and TextWorld), the learned memory designs tend to store fine-grained knowledge, such as spatial relationships between objects and room layouts. In contrast, for tasks that require more complex reasoning (e.g., Baba Is AI and MiniHack), the memory designs favor domain-specific abstract strategy, including strategy libraries and plan synthesis. This pattern indicates that \ouralgo automatically specializes memory designs to the demands of each domain.

\begin{figure}[t]
  \begin{center}
    \centerline{\includegraphics[width=0.9\columnwidth]{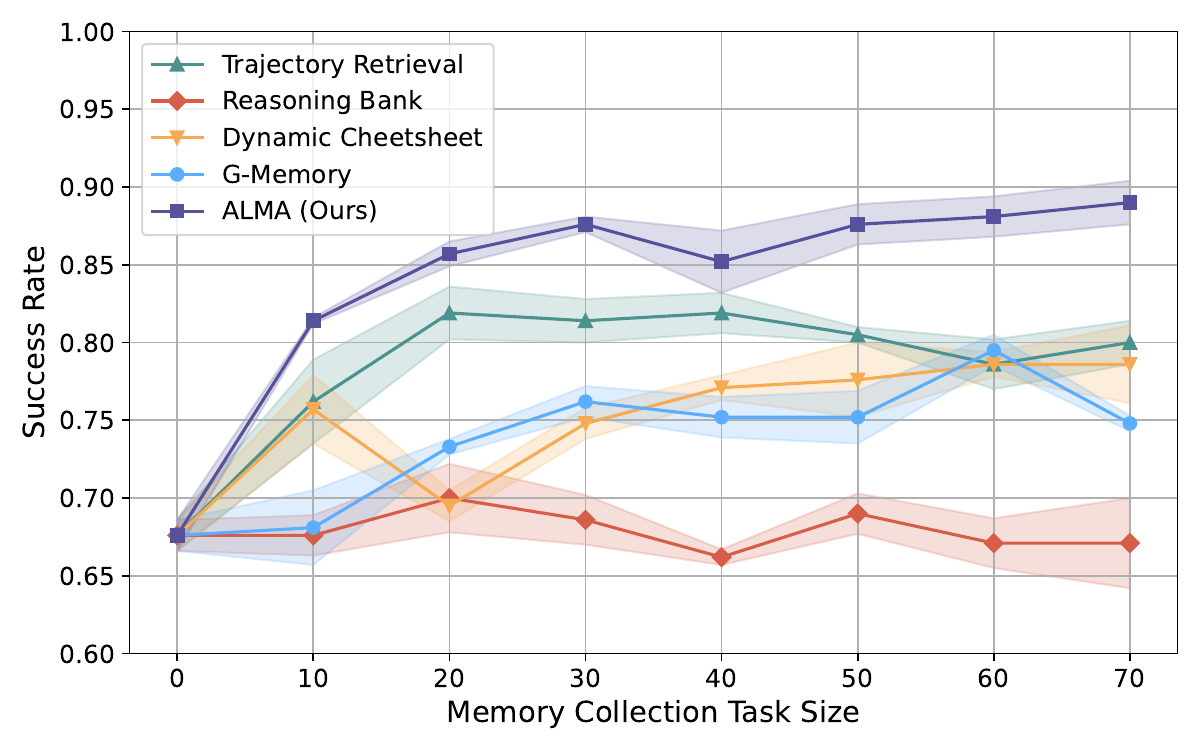}}
    \captionsetup{aboveskip=5pt, belowskip=-2pt}
    \caption{
        \textbf{Success rates of different memory designs in ALFWorld as the task size during the memory collection phase increases.} Evaluations are performed using static mode during the Deployment Phase to study how performance scales with collected static memory. \texttt{GPT-5-mini} is used as the FM in the agentic system during testing. Shaded areas indicate standard error, calculated over three runs of the Deployment Phase. The learned memory design achieves higher performance faster with limited data and scales better than human-designed baselines. 
    }
    \label{scale_plot}
  \end{center}
\end{figure}
To study how performance scales with experience, we evaluate memory designs using different sizes of tasks during the Memory Collection Phase and using static mode during the Deployment Phase.
As shown in \cref{scale_plot}, compared to manual memory designs, the learned memory design achieves higher performance faster with limited data and scales better as more trajectories are provided, demonstrating both higher sample efficiency and stronger scalability.
We observe that increasing the trajectory size does not necessarily lead to a monotonic improvement in success rate, which is expected since trajectories may vary in quality \citep{ouyang2025reasoningbank}.

\begin{figure}[t]
  \begin{center}
    \centerline{\includegraphics[width=0.9\columnwidth]{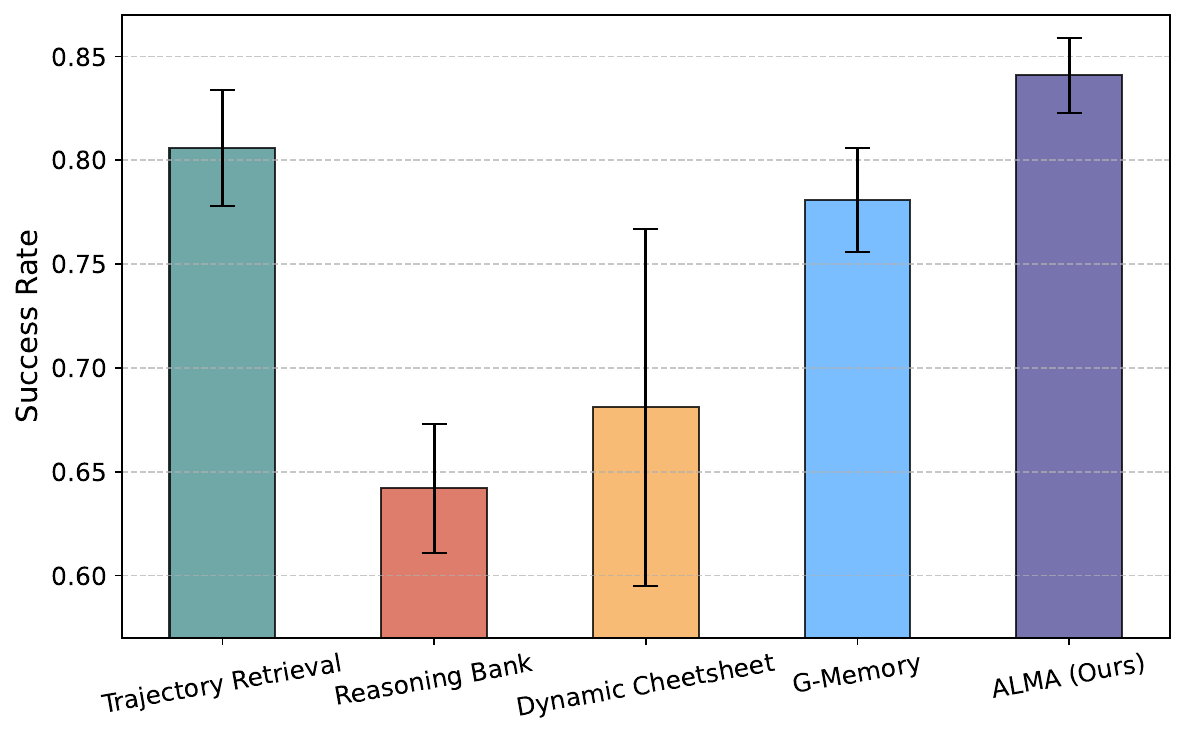}}
        \captionsetup{
      aboveskip=5pt,  
      belowskip=-2pt  
    }
    \caption{
    \textbf{Success rates of different memory designs in ALFWorld under task distribution shift in the Deployment Phase.} Memory is collected from tasks in \textit{valid\_seen} during the Memory Collection Phase and evaluated on the \textit{valid\_unseen} dataset with dynamic mode during the Deployment Phase, using \texttt{GPT-5-mini} as the FM in the agentic system. The error bars indicate the standard errors calculated over three runs of the Deployment Phase. The learned memory design adapts more effectively than human-designed baselines under task distribution shift. 
    }
    \label{trans_plot}
  \end{center}
\end{figure}
We evaluate the adaptability of learned memory designs to the distribution shift of tasks.
To evaluate adaptation under distribution shift, memory is collected from 70 tasks in the \textit{valid\_seen} during the Memory Collection Phase and evaluated on the \textit{valid\_unseen} using dynamic mode during the Deployment Phase (Appendix~\ref{app:testing_details}). This tests whether memory designs can adapt to tasks with distribution shift in \textit{valid\_unseen} by dynamically updating memory. 
As shown in \cref{trans_plot}, the learned memory design achieves a success rate of 84.1\% on ALFWorld, outperforming all human-designed baselines and demonstrating more effective adaptation under task distribution shift. 

\begin{figure}[t]
  \begin{center}
    \centerline{\includegraphics[width=0.9\columnwidth]
    {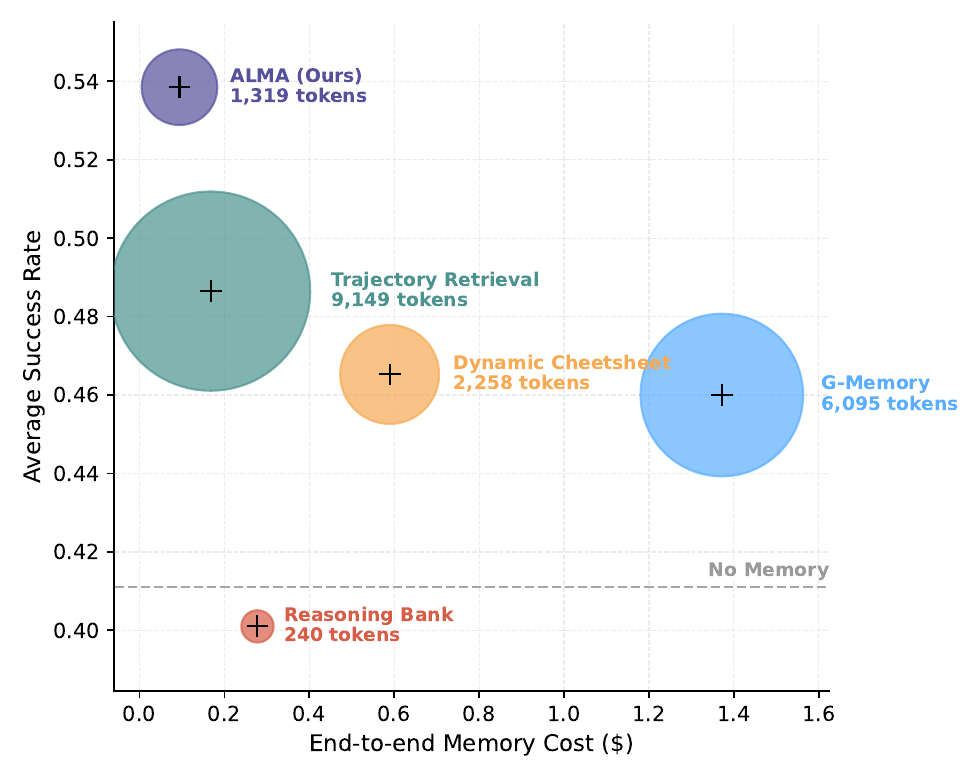}}
    \captionsetup{
      aboveskip=2pt,  
      belowskip=-2pt  
    }
    \caption{
    \textbf{Cost efficiency versus task performance across all benchmarks for different memory designs.} The x-axis shows the overall end-to-end memory cost across benchmarks. Bubble size represents the average token size of retrieved content inserted into the agentic system's context per task. The learned memory design outperforms all baselines while being cost-efficient. 
    }
    \label{tradeoff_combined}
  \end{center}
\end{figure}

We measure the cost efficiency of learned memory designs by studying their end-to-end memory cost, defined as the total cost of FMs required to transform raw interaction logs into retrieved content inserted into the agentic system's context across the Memory Collection and Deployment Phases. We also include the token size of the retrieved content. The details are available in Appendix~\ref{eval_cost}.
As shown in \cref{tradeoff_combined}, compared to all manual designs, our learned memory designs achieve an average success rate of 53.9\%, with lower overall end-to-end memory cost across all benchmarks (\$0.09) and appropriate token sizes of retrieved content. The results show \ouralgo effectively balances memory efficiency and task performance. Since the current algorithm does not explicitly optimize for cost efficiency, we anticipate that incorporating techniques such as multi-objective optimization in future work could enable even more effective exploration of the efficiency-performance Pareto front.


\vspace{-0.6em}
\section{Conclusion, Safety, and Future Work}
We propose \ouralgo, a paradigm that learns to discover new memory designs, enabling agentic systems to continually learn from their past experiences. We evaluate \ouralgo on a range of sequential decision-making benchmarks and the best-learned designs consistently outperform manually designed baselines, demonstrating the effectiveness of our approach. Furthermore, subsequent experiments show that the learned memory designs exhibit strong scalability, transferability, and cost efficiency. Overall, \ouralgo provides a principled approach to learn memory design, taking a step toward enabling continual learning in agentic systems.


\ouralgo represents a step toward continual learning in AI, contributing to the broader goal of developing AI-generating algorithms beyond current manual approaches~\citep{clune2019ai}. However, such AI-generating algorithms introduce unique safety concerns, as components of AI systems are learned rather than designed by humans. The learned components may inadvertently introduce behaviors that deviate from human intentions, ignoring optimization guided by a predefined target metric~\citep{Bostrom2020EthicalAI, Md2020SourceFinder, Ecoffet2020Open}. 
Recognizing these challenges, we impose explicit constraints and restrictions in our experimental setup during the learning process. 
Specifically, each memory design is generated as code by the Meta Agent, they are validated and executed within isolated sandbox environments, with access confined to the sandbox to prevent any interference with the external system, mitigating the risk of unintended actions. We also perform human oversight of learned memory designs to ensure that no potentially harmful actions (e.g., prompt injection) are included in memory designs. In future work, it will be essential to establish a systematic inspection mechanism as the system scales and is deployed, which may include AI and human inspection.

While \ouralgo demonstrates strong effectiveness in learning memory designs, it has several limitations that open up future research directions. One limitation is that the memory designs are learned using a pre-defined learning set, instead of dynamically learning memory designs when facing new tasks. Ideally, an adaptive continual learner would be able to learn memory designs online from dynamic data, without separating learning and testing phases. While \ouralgo has the potential to support online learning, we do not pursue this approach due to computational budget constraints, as evaluating each explored memory design would require a large number of rollouts. Therefore, a possible future direction is to enable online learning of memory design that continually adapts to arbitrary domains.
Another limitation stems from the fact that we demonstrate learning memory designs in code space. While effective, their capabilities may be limited by the underlying FMs. 
Future work could explore automated designing and training novel FM architectures with native memory support.

Overall, \ouralgo represents a step forward toward automated, continual-learning AI. While current limitations prevent us from demonstrating a system that learns both memory and the agentic system, our results highlight its potential to surpass manually designed memory, showing a way towards ``learning to continually learn''.
When developed with the appropriate safety measures, \ouralgo holds promise for a general-purpose AI capable of learning to continually learn across domains and autonomously adapting to new ones.

\section*{Impact Statement}
This paper introduces \ouralgo, a paradigm that automatically learns novel memory designs, thereby enabling agentic systems to continually learn. By leveraging a Meta Agent with open-ended exploration, \ouralgo mitigates the intensive manual effort conventionally required to handcrafted memory designs for diverse domains. Furthermore, our findings highlight a path toward agentic systems capable of tailoring their memory designs to any context, fostering the development of continual learning of AI. These insights empower practitioners to: (1) automate the development of domain-specific memory designs for specialized fields such as medicine, finance, and software engineering; and (2) uncover memory paradigms that have yet to be identified through human intuition. Ultimately, this work advances the automation of memory design for agentic systems and serves as a foundation for future research on learning memory designs.

\section*{Acknowledgments}
This research was supported by the Vector Institute, the Canada CIFAR AI Chairs program, a grant from Schmidt Futures, an NSERC Discovery Grant, and a generous donation from Rafael Cosman. Resources used in preparing this research were provided, in part, by the Province of Ontario, the Government of Canada through CIFAR, and companies sponsoring the Vector Institute (\url{https://vectorinstitute.ai/partnerships/current-partners/}). Any opinions, findings, and conclusions or recommendations expressed in this material are those of the authors and do not necessarily reflect the views of the sponsors.

\nocite{langley00}

\bibliography{example_paper}
\bibliographystyle{icml2026}

\newpage
\appendix
\onecolumn

\section{Algorithmic Details}
\subsection{Search Space}
\label{search_space}
\begin{lstlisting}[
  style=frame,
  language=Python,
  caption={The Python abstract class used in the learning process. In the code, sub-modules are implemented as layers, with each layer encapsulating the functionality of a module. The overall execution order of these update and retrieval operations across sub-modules is orchestrated by the general interfaces: \texttt{general\_retrieve} and \texttt{general\_update}. The example of learned memory design is available at Appendix ~\ref{app:learned_design}
},
  xleftmargin=0.025\columnwidth, % 3. 左边留 2.5% 的空隙
  xrightmargin=0.025\columnwidth, % 4. 右边留 2.5% 的空隙
  linewidth=0.95\columnwidth % 5. 总宽度 95%，配合左右 margin 实现居中
]
from abc import ABC, abstractmethod
from typing import Dict, Optional, Any
from dataclasses import dataclass
from eval_envs.base_envs import Basic_Recorder

@dataclass
class Sub_memo_layer(ABC):
    """
    Abstract class for retrieve/update sub-function
    """
    layer_intro: str = "Introduction of the structure of current defined Database(if any), corresponding Update and Retrieve method."
    database: Optional[Any] = None

    @abstractmethod
    async def retrieve(self, **kwargs):
        """
        The retrieve function of current layer. 
        """
        pass

    @abstractmethod
    async def update(self, **kwargs):
        """
        The update function of current layer. 
        """
        pass

class MemoStructure(ABC):

    def __init__(self):
        self.database: Optional[Any] = None
        
    # -------- Pipeline Runner --------
    @abstractmethod
    async def general_retrieve(self, recorder: Basic_Recorder) -> Dict: 
        """
        The general retrieve method, use the retrieve function in each layer, determine order and input-output usage by yourself.
        """
        pass

    @abstractmethod
    async def general_update(self, recorder: Basic_Recorder) -> None:  
        """
        The general update method, use the update function in each layer, determine order and input-output usage by yourself.
        """
        pass
\end{lstlisting}

\subsection{Evaluation of Memory design}
\label{eval_process}
We first formally define a memory design as a triplet $\mathcal{M} = (U, D, R)$, where $U$ and $R$ denote the update and retrieval processes that interface with the agentic system, and $D$ represents the internal storage structure used for memory retention.
To evaluate a memory design on a dataset $\mathcal{G}$, 
we split $\mathcal{G}$ evenly into two subsets: 
a set $\mathcal{G}_{\text{collection}}$ for memory collection  and a set $\mathcal{G}_{\text{deployment}}$ for evaluation during the Deployment Phase. 
During the Memory Collection Phase, the agentic system interacts with the environment to finish tasks in $\mathcal{G}_{\text{collection}}$ without any memory access, 
producing interaction logs that are used to perform memory updates and construct a static memory.
During the Deployment Phase, the agentic system retrieves relevant contexts from the static memory, which are then integrated into the agentic system to facilitate task execution within $\mathcal{G}_{\text{deployment}}$.
The performance on $\mathcal{G}_{\text{deployment}}$ reflects the ability of a memory design to collect and utilize knowledge.

Furthermore, when evaluated on a dataset with task distribution shift during the Deployment Phase, the agentic system first retrieves existing knowledge to construct the knowledge context for the new task, and subsequently updates the memory with the resulting interaction log.

\paragraph{Memory Collection Phase.}
$\mathcal{G}_{\text{collection}}$ is used to collect a set of interaction trajectories 
$\mathcal{T}_{\text{collection}} = \{ \tau^{(j)}_{\text{collection}} \}_{j=1}^{N}$ 
without any memory access. 
Each trajectory $\tau^{(j)}_{\text{collection}}$ takes the form
\[
\tau^{(j)}_{\text{collection}} 
= \big( s^{(j)}_1, a^{(j)}_1, s^{(j)}_2, a^{(j)}_2, \ldots, s^{(j)}_{n_j}, a^{(j)}_{n_j} \big),
\quad 
a^{(j)}_t \sim \pi(\cdot \mid s^{(j)}_{\le t}, a^{(j)}_{<t})
\]
where $s$ denotes the environment state, and $a$ denotes the action produced by the fixed agentic policy $\pi$.
For each trajectory, we further obtain a task-level feedback
\[
f^{(j)}_{\text{collection}} = F\!\left(\tau^{(j)}_{\text{collection}}\right)
\]
where $F(\cdot)$ denotes the benchmark-specific evaluation function that assigns a scalar score to each trajectory. We then sequentially update the memory database using the collected interaction logs
\[
D_j = U\!\left(\tau^{(j)}_{\text{collection}}, f^{(j)}_{\text{collection}}, D_{j-1}\right)
\]
where the memory state is updated after each collected trajectory. The final collected memory is denoted as $D_N$.

\paragraph{Deployment Phase.} 
During the Deployment Phase, we evaluate the performance of the agentic system, with knowledge retrieved from $D_N$, 
on the deployment set $\mathcal{G}_{\text{deployment}} = \{ g^{(i)}_{\text{deployment}} \}_{i=1}^K$.

For each task $g^{(i)}_{\text{deployment}}$, we first retrieve relevant knowledge from the memory state $D_N$
\[
e^{(i)} = R(D_N, s^{(i)}_{1})
\]
where $s^{(i)}_{1}$ denotes the initial state, including observation and task objective of the current task. 
The agent then interacts with the environment, resulting in a trajectory $\tau^{(i)}_{\text{deployment}}$, where each action is sampled according to
\[
a^{(i)}_t \sim \pi(\cdot \mid s^{(i)}_{\le t}, a^{(i)}_{<t}, e^{(i)})
\]
The feedback is calculated as $f^{(i)}_{\text{deployment}} = F(\tau^{(i)}_{\text{deployment}})$. 
The overall performance of the memory design $\mathcal{M}$ is then computed as the average over all deployment tasks
\[
f_{\mathcal{M}} = \frac{1}{K}\sum_{i=1}^K f^{(i)}_{\text{deployment}}
\]
When performing Deployment Phase with distribution shift tasks, memory state is further updated sequentially using the collected trajectories $\tau^{(i)}_{\text{deployment}}$ 
and feedback signals $f^{(i)}_{\text{deployment}}$, producing $D_{N+i}$. 
The updated memory state is then used for retrieval when performing the $(i+1)$-th task.

\subsection{Evaluation Cost}
\label{eval_cost}
\paragraph{End-to-End Memory Cost.} This metric quantifies the cumulative computational overhead incurred by the FMs to generate the content integrated into the agentic system with static mode in Deployment Phase. The cost includes: (1) the expense during the Memory Collection Phase to produce the memory state $D_N$; and (2) the cumulative cost of generating knowledge $\{e^{(i)}\}_{i=1}^K$ for all tasks during the Deployment Phase. We suggest that this provides a fair and unbiased way to measure the cost of a memory design, independent of its effectiveness. During the Memory Collection Phase, the agentic system interacts with environments without memory access, ensuring that all trajectories are unaffected by the memory design’s performance. Since these trajectories are the only input used to produce memory, the costs of producing the memory state $D_N$ and generating relevant knowledge during static deployment depend solely on these trajectories and are therefore independent of the memory design’s effectiveness. We present a summary of the costs associated with testing a memory design across all benchmarks, including \texttt{GPT-4o-mini} and \texttt{text-embedding-3-small}. 

\paragraph{Token Size of Retrieved Knowledge.} We measure the token count of the retrieved context $e^{(i)}$ for each task during the static Deployment Phase. This metric quantifies the informational overhead injected into the agent's prompt, serving as a key proxy for measuring the inference-time efficiency of the memory design $\mathcal{M}$. The token size is also independent of effectiveness of a memory design, since it is dependent solely on trajectories collected in Memory Collection Phase. We utilize the \texttt{GPT-5-mini} tokenizer to calculate the token sizes.

\subsection{Memory Design Archive and Sampling Process}
\label{selection_process}
\paragraph{Memory Design Archive.} 
To iteratively explore new memory designs, we maintain a memory design archive $\mathcal{A}_l$ that stores all designs discovered up to learning step $l$. For each candidate design $\mathcal{M}$, we perform a stratified sampling process \citep{thompson2012sampling} after its evaluation on a large deployment set. Specifically, instead of storing the entire interaction logs, we extract a small subset of size $K_f$ as a proxy for the design's performance. This subset is denoted as
\[
\mathcal{S}_{\mathcal{M}} = \{ (e^{(i)}, \tau^{(i)}_{\text{deployment}}, f_{\text{deployment}}^{(i)}) \}_{i=1}^{K_f}
\]
where each tuple contains the retrieved context $e^{(i)}$, the resulting trajectory $\tau^{(i)}_{\text{deployment}}$, and the task-level feedback $f_{\text{deployment}}^{(i)}$. 

The archive further records the overall performance $f_{\mathcal{M}}$ and the times a memory design is being sampled $t_{\mathcal{M}}$, which tracks the number of times $\mathcal{M}$ has been sampled and evaluated across $l$ learning steps. Formally, at step $l$, the memory design archive is defined as the collection of these records
\[
\mathcal{A}_l = \Big\{ (\mathcal{M}, f_{\mathcal{M}}, \mathcal{S}_{\mathcal{M}}, t_{\mathcal{M}}) \;\Big|\; \mathcal{M} \text{ discovered before step } l \Big\}.
\]

\paragraph{Sampling Process.} Similar to prior exploration approaches \citep{ecoffet2021first, zhang2025darwin}, given the current memory design archive, we perform a sampling process to sample memory designs that will be used in the next learning step. 
For each memory design $\mathcal{M}_i$, we first compute the normalized performance relative to a baseline $f_0$
\[
\hat f_{\mathcal{M}_i} = \sigma(f_{\mathcal{M}_i} - f_0) = \frac{1}{1 + \exp(-\lambda (f_{\mathcal{M}_i} - f_0))}
\]
where $f_0$ denotes the performance of the agentic system on $\mathcal{G}_{\text{deployment}}$ without memory access. 
Subtracting $f_0$ ensures that the normalized score reflects the performance gain provided by the memory design.
The final sampling score of each memory design is then given by
\[
J_i \equiv J_{\mathcal{M}_i} = \hat f_{\mathcal{M}_i} - \alpha \log(1 + t_{\mathcal{M}_i})
\]
where $t_{\mathcal{M}_i}$ is the number of times $\mathcal{M}_i$ has been sampled, and $\alpha$ controls the strength of the penalty. 
We then compute the sampling probability using a softmax over the final sampling scores
\[
p_i = \frac{\exp\big(J_i / T - \max_j J_j / T\big)}{\sum_{j=1}^{|\mathcal{A}_l|} \exp\big(J_j / T - \max_j J_j / T\big)} , \quad i = 1, \dots, |\mathcal{A}_l|
\]
where $T$ is a temperature parameter controlling the smoothness of the distribution. The sampling mechanism prioritizes high-performing yet under-sampled memory designs to balance exploration and exploitation. Meanwhile, designs with moderate performances retain a non-zero sampling probability to maintain diversity throughout the learning process.
Finally, we sample memory designs according to $\{p_i\}$:
\[
\mathcal{M}_{\text{sampled}} \subset \{\mathcal{M}_1, \dots, \mathcal{M}_{|\mathcal{A}_l|}\}, \quad 
\mathcal{M}_{\text{sampled}} \sim \text{Categorical}(p_1, \dots, p_{|\mathcal{A}_l|})
\]
\subsection{Prompts of Meta Agent}
\label{prompts_meta}

\begin{topblueprompt}{Prompt for Ideate \& Planning}
\begin{lstlisting}[
  style=clean]
You are a **Senior Agent Construction Engineer** responsible for provide suggestions for a memory structure written by a entry level engineer, to make the memory structure better for downstream agent to finish tasks.
### Memo Information Overview
1. **source_code**
    - Each layer(which inherits `Sub_memo_layer`) contains:
      - **Retrieve**: fetches relevant memory elements from the database.
      - **Update**: writes or modifies entries in the database.
    - Final MemoStructure(which inherits `MemoStructure`) contains:
        - **general_retrieve**: general retrieve method, that contain the order and input-output usage of retrieve function of each layer.
        - **general_update**: general update method, that contain the order and input-output usage of update function of each layer.
    - code usage: Your memory structure will be used in the agent workflow:
        - `general_retrieve(recorder)`: used **before** start executing the task, to retreive task relevant information. Your output json will be directly send to agent, so please make sure your output is well organized, include all useful information and avoid redundency, in a agent understandable way.
        - `general_update(recorder)`: used **after** task is finished, to update the trajectory, reward, or other information.
2. **examples**
    - **examples**: some trajectries, including the memory retrieved(memory retrived should only happened before starting the task), and the steps of actions took by agent.
3. **benchmark_eval_score**
    -  performance(success rate) of current memory stucture + general agent system. Need to use the score to analyze the performance and bottleneck of current memory structure.

### Your Task:
You will analyze past suggestion examples(including past source code, suggestions, and the improve score it led to) and the current retrieved trajectories and memory source code, then produce concrete, prioritized suggestions to improve the memory structure.
Follow the numbered procedure below and produce the requested structured outputs.
Step 1 - Learn from past suggestions & the improve score
    1. Look at the provided improve_score (positive as improvement, negative as degradation) and the single suggestion_example that produced that score.
    2. Explain why that suggestion led to improvement or degradation:
        - What pattern in the change made it succeed or fail?
        - Which behaviors, assumptions, or shortcuts in that suggestion were helpful? Which were harmful?
        - From these concrete cases, extract 2 to 5 general principles to adopt and 2 to 5 pitfalls to avoid when creating future suggestions.

Step 2 - Inspect sampled trajectories and benchmark performance and decide which memories are useful
    1. Review the current trajectory_examples (they include episodes with varying rewards).
    2. For each retrieved memory item (or memory group) returned for the trajectory, label it as one of:
        - Useful & Relevant - clearly applies to the current situation and can guide action;
        - Potentially Useful - has value but needs reformatting, summarization, or indexing to be helpful;
        - Irrelevant / Confusing - not related to this trajectory or misleading;
        - Empty / Badly Formatted - blank, placeholder, or not parseable. 
    3. For each memory you mark Useful/Potentially Useful, say how it would help (e.g., provides a repeated subgoal, highlights a trap, identifies key object interactions).
    4. For Irrelevant/Empty items, explain why they failed retrieval combine with the memory source code (e.g., wrong keying, over-specific content, missing summarization).

Step 3 - Inspect memory source and produce concrete suggestions
    1. Review the memory source code (retrieval keys, indexing, storage format, layers). Using Step 1 principles and Step 2 labels, propose specific changes to the memory system that address the observed issues.
    2. Combined the memory source code with your analysis in step 2, giving suggestions. For each suggested change, include:
        - What to change (code-level or pipeline change, e.g., add summarization layer, change indexing key, normalize objects to noun-phrases).
        - Why it will help (link back to a principle or a concrete failing you observed).
    3. Prioritize suggestions: label them High / Medium / Low priority and give an implementation order.
    4. Link Analysis to Benchmark Performance
        - Use benchmark_eval_score to identify which structural weaknesses correlate with poor performance.

Extra checks (quality & coherence)
    1. Flag obvious content issues: duplicates, empty entries, raw dumps, mis-typed fields, or numeric types that break JSON serialization.
    2. Check layer interaction: do layers pass structured outputs to each other, or only dump free-form text?
    3. If retrieval returns empty lists or dicts, emphasize structural fixes (keying, ensure type consistency, avoid over-relying on try/except fallbacks).

Goal: Combine reflection on past improvement signals with current system diagnosis to produce actionable, high-level suggestions that strengthen memory structure quality.

### Benchmark Information:
{TASK_DESCRIPTION[task_type]}

### Required Output:
Return a JSON object followed below json schema:
{json.dumps(MEMO_ANALYSIS_OUTPUT_FORMAT, indent = 2, ensure_ascii = False)}


\end{lstlisting}
\end{topblueprompt}
\begin{topblueprompt}{Prompt for Implementation}
\begin{lstlisting}[
  style=clean]
You are a senior AI software engineer. Your task is to build an agent memory system composed of multiple specialized memory layers and a coordinating memory structure. The agent will be used in {task_type}. Yout memory structure aims to help down stream agent to have more relevant advice, experience to help it further finish current task.
    
{TASK_DESCRIPTION[task_type]}

You are given the following two base classes:
<BACKBONE_CODE>
{basic_classes}
</BACKBONE_CODE>
Inherit these two base classes and recorder by writing:
```python
from agents.memo_structure import Sub_memo_layer, MemoStructure
from eval_envs.base_envs import Basic_Recorder
from utils.hire_agent import Agent, Embedding
from langchain_chroma import Chroma
```

<CODE_INPUT>
{interaction_prompt}
</CODE_INPUT>

<CODE_USAGE>
Your memory structure will be used in the agent workflow:
    - `general_retrieve(recorder)`: used **before** start executing the task, to retreive task relevant information. Your output json will be directly send to agent, so please make sure your output is well organized, include all useful information and avoid redundency, in a agent understandable way.
    - `general_update(recorder)`: used **after** task is finished, to update the trajectory, reward, or other information.
<CODE_USAGE>

Here is the basic tools provided:
<GRAPH_DATABASE_INTERACTION>
{NXGRAPH_CHEETSHEET}
</GRAPH_DATABASE_INTERACTION>

<CHROMA_DATABASE_INTERACTION>
{CHROMA_CHEETSHEET}
</CHROMA_DATABASE_INTERACTION>

<OTHER_TOOLS>
{TOOL_CHEETSHEET}
</OTHER_TOOLS>

### Your Task:
Modify the code above so that it fully satisfies the following design goals:

1. **Multiple Memory Layers:**   
   - Create multiple subclasses of `Sub_memo_layer`.
   - Each layer must have a **clear responsibility** and you might need to maintain its own database(based on nx.graph or chroma).  
   - Think carefully about **what type of data** belongs in each layer's database.

2. **General Retrieve/Update Orchestration:**  
   - Create a subclass of `MemoStructure` that orchestrates all layers.  
   - `general_retrieve()` should intelligently chain the results:  
     - Output from layer 1 can become input to layer 2.  
     - You should design a reasonable order of retrieval.  
   - `general_update()` should propagate updates to relevant layers in a sensible order.  
   - You are free to reorder calls or preprocess inputs to achieve coherent memory behavior.

3. **Out-of-the-Box Reasoning:**  
   - Do not just mechanically call each layer one by one - think about the **semantic flow of information**.  
   - Consider cases like:  
     - what type of memory layer can be used according to the analysis result or task description, with the aim to better assist the agent to finish it's task? 
     - what order and input output should be best suitable for the analysis result or task description? 
     - Think about high-level stategy: what can be a good memory structure, and has good ability to transfer to other area? 
   - Make sure each layer plays a meaningful role in the system.
   - Directly perform simple plans or content writing based on if/else patterns should be avoided, since this will hurt the transfer ability.
   - Keep the retrieved memory clean and useful, aviod cutting off meaningful texts, repeat same patterns in the retrieved memory.

4. **Integration with Utilities:**  
   - Feel free to use any provided utility functions (e.g., similarity calculation, interaction with databases, hire new agent) if relevant. The tools available will be listed in `TOOLS` section.
   - You can also create your own tools if neccessary, think out of the box.

5. **Code Quality:**  
   - Output clean, runnable Python code following PEP8.  
   - Ensure `general_retrieve()` and `general_update()` accept `Basic_Recorder` and orchestrate the pipeline end-to-end.  
   - Initialize all layers in `MemoStructure.__init__`.
   - Do not overuse defensive programming; raise appropriate exceptions when unexpected conditions occur to facilitate debugging.

6. **Coherent Policy Logic:**
    - Avoid placeholders like pass or # TODO.
    - Avoid hard-coded if/else branches or enumerated case handling; instead, express the logic through modular policy functions, scoring mechanisms, or composable decision rules.
    - Instead of enumerating case-specific rules, express generalizable principles that could apply across different families or new unseen tasks.
    - The logic should be adaptable and compositional, not dependent on predefined constants or string names.
    - Use abstractions instead of specific family identifiers.

The goal is to ensure the memory policy behaves consistently across tasks and supports generalization, not to hard-code specific task behaviors.

### Important:
- Think creatively about data flow - outputs of one layer can feed into the next.
- Each layer's functionality and stored data should be clearly designed.
- Provide **only the final rewritten code**, no explanations.
\end{lstlisting}
\end{topblueprompt}
\begin{topblueprompt}{Prompt for Debugging}
\begin{lstlisting}[
  style=clean]
You are a senior AI software engineer and code repair expert.
Your role is to carefully analyze the provided code and the error information, identify potential errors or design flaws, and directly rewrite or edit the code to fix those issues - while keeping the main design goals and intentions exactly the same.

You are given the following context and base classes:
<BACKBONE_CODE>
{basic_classes}
</BACKBONE_CODE>

<CODE_INPUT>
{interaction_prompt}
</CODE_INPUT>

<CODE_USAGE>
Your memory structure will be used in the agent workflow:
    - `general_retrieve(recorder)`: used **before** start executing the task, to retreive task relevant information. Your output json will be directly send to agent, so please make sure your output is well organized, include all useful information and avoid redundency, in a agent understandable way.
    - `general_update(recorder)`: used **after** task is finished, to update the trajectory, reward, or other information.
<CODE_USAGE>

### Your Task:
Carefully inspect the code, and error information, detect the root cause of the errors, structural issues, or missing implementations, and **only fix the root cause code**. Here are some cheetsheet that could be useful:

<GRAPH_DATABASE_INTERACTION>
{NXGRAPH_CHEETSHEET}
</GRAPH_DATABASE_INTERACTION>

<CHROMA_DATABASE_INTERACTION>
{CHROMA_CHEETSHEET}
</CHROMA_DATABASE_INTERACTION>

<OTHER_TOOLS>
{TOOL_CHEETSHEET}
</OTHER_TOOLS>

### Output:
Return **only the final corrected Python code**, no explanations or commentary.
\end{lstlisting}
\end{topblueprompt}

\subsection{Pseudocode}
\label{Pseudocode}

\begin{algorithm}[H]
\caption{Learning of Memory Designs}
\label{alg:meta_learning}
\begin{algorithmic}[1]

\STATE {\bfseries Input:} 
initial archive $\mathcal{A}_0 = \emptyset$; 
meta-learning steps $L$; 
collection tasks $\mathcal{G}_{\text{collection}}$; 
deployment tasks $\mathcal{G}_{\text{deployment}}$.

\STATE {\bfseries Output:} 
optimal memory design $\mathcal{M}^*$ and final archive $\mathcal{A}_L$.

\FOR{$l = 1$ {\bfseries to} $L$}

    \item[] \textit{// Sampling Process.}

    \STATE Sample a set of base memory designs
    $
    \mathcal{M}_{\text{sampled}} \subset \mathcal{A}_{l-1}
    $
    based on historical performance and visit counts.
    \STATE Update visit time for each sampled design:
    $
    t_{\mathcal{M}_{\text{base}}} \leftarrow t_{\mathcal{M}_{\text{base}}} + 1, \quad \forall \mathcal{M_{\text{base}}} \in \mathcal{M}_{\text{sampled}}
    $
    \item[] \textit{// Parallel Refinement and Evaluation.}
    \FORALL{$\mathcal{M}_{\text{base}} \in \mathcal{M}_{\text{sampled}}$ {\bfseries in parallel}}

        \STATE $\rho \leftarrow \textsc{Planner}(\mathcal{M}_{base},\mathcal{S}_{\mathcal{M}_{\text{base}}}, f_{\mathcal{M}_{base}})$
        \hfill \textit{// Planning}

        \STATE
        $\tilde{\mathcal{M}} \leftarrow \textsc{Implementor}(\mathcal{M}_{\text{base}}, \rho)$
        \hfill \textit{// Implementation}
        \STATE $error\_detected \leftarrow \textsc{TrialRun}(\tilde{\mathcal{M}})$
        \IF{$error\_detected$}
            \FOR{$i = 1, \dots, N$}  
                \STATE $\tilde{\mathcal{M}} \leftarrow \textsc{Debugger}(\tilde{\mathcal{M}}, error\_detected)$
                \hfill \textit{// Retry attempt $i$}
                \STATE $error\_detected \leftarrow \textsc{TrialRun}(\tilde{\mathcal{M}})$
                \IF {not $error\_detected$} \STATE \textbf{break} \ENDIF
            \ENDFOR
        \ENDIF
        
        \item[] \textit{// Memory Collection Phase.}
        \STATE Initialize $D_0 \leftarrow \emptyset$
        \FORALL{$g^{(j)} \in \mathcal{G}_{\text{collection}}, j=1,\dots,|\mathcal{G}_{\text{collection}}|$}
            \STATE $\tau_{g^{(j)}} 
= \big( s_1, a_1, s_2, a_2, \ldots, s_{n_{g^{(j)}} }, a_{n_{g^{(j)}} } \big),
\quad 
a_t \sim \pi(\cdot \mid s_{\le t}, a_{<t})$
            \hfill \textit{// memory-free trajectory collection}
            \STATE $f_{g^{(j)}} \leftarrow F(\tau_{g^{(j)}})$
            \hfill \textit{// Feedback calculation}
            \STATE
            $
            D_j \leftarrow U_{\tilde{\mathcal{M}}}(\tau_{g^{(j)}}, f_{g^{(j)}}, D_{j-1})
            $
            \hfill \textit{// Memory update }
        \ENDFOR

        \item[] \textit{// Deployment Evaluation Phase.}
        \FORALL{$g \in \mathcal{G}_{\text{deployment}}, k=1,\dots,|\mathcal{G}_{\text{deployment}}|$}
            \STATE $e_{g^{(k)}} \leftarrow R_{\tilde{\mathcal{M}}}(D_{|\mathcal{G}_{\text{collection}}|}, s_{g^{(k)}})$
            \hfill \textit{// Retrieve experience }
            \STATE $\tau_{g^{(k)}}= \big( s_1, a_1, s_2, a_2, \ldots, s_{n_{g^{(k)}}}, a_{n_{g^{(k)}}} \big),
\quad 
a_t \sim \pi(\cdot \mid s_{\le t}, a_{<t}, e_{g^{(k)}})$ 
            \hfill \textit{// Evaluation with memory}
            \STATE $f_{g^{(k)}} \leftarrow F(\tau_{g^{(k)}})$
            \hfill \textit{// Feedback calculation}
        \ENDFOR

        \STATE Aggregate deployment performance $f_{\tilde{\mathcal{M}}} \leftarrow \frac{1}{K}\sum f_{g^{(k)}}$

        \STATE Store evaluation samples
        $
        \mathcal{S}_{\tilde{\mathcal{M}}} \leftarrow \{ (e_g, \tau_g, f_g) \}
        $

        \STATE Insert
        $
        (\tilde{\mathcal{M}}, f_{\tilde{\mathcal{M}}}, \mathcal{S}_{\tilde{\mathcal{M}}}, t_{\tilde{\mathcal{M}}})
        $
        into $\mathcal{A}_l$
        \hfill \textit{// Archive update}

    \ENDFOR

\ENDFOR

\STATE
$
\mathcal{M}^* \leftarrow \arg\max_{\mathcal{M} \in \mathcal{A}_L} f_{\mathcal{M}}
$
\hfill \textit{// Output optimal memory design}

\end{algorithmic}
\end{algorithm}

\section{Experiment Details}
\label{exp_details}
\subsection{Benchmark Details}
\label{app:benchmark}
We evaluate \ouralgo on four sequential decision-making benchmarks: ALFWorld \citep{shridhar2021Alfworld}, TextWorld \citep{textworld}, Baba Is AI \citep{cloos2024baba}, and MiniHack \citep{samvelyan2021minihack}. 
All benchmarks involve multi-step interactions, providing a suitable testbed for evaluating how agents use memory designs to learn from experience.

\begin{figure}[htbp]
  \centering
  \includegraphics[width=1\textwidth]{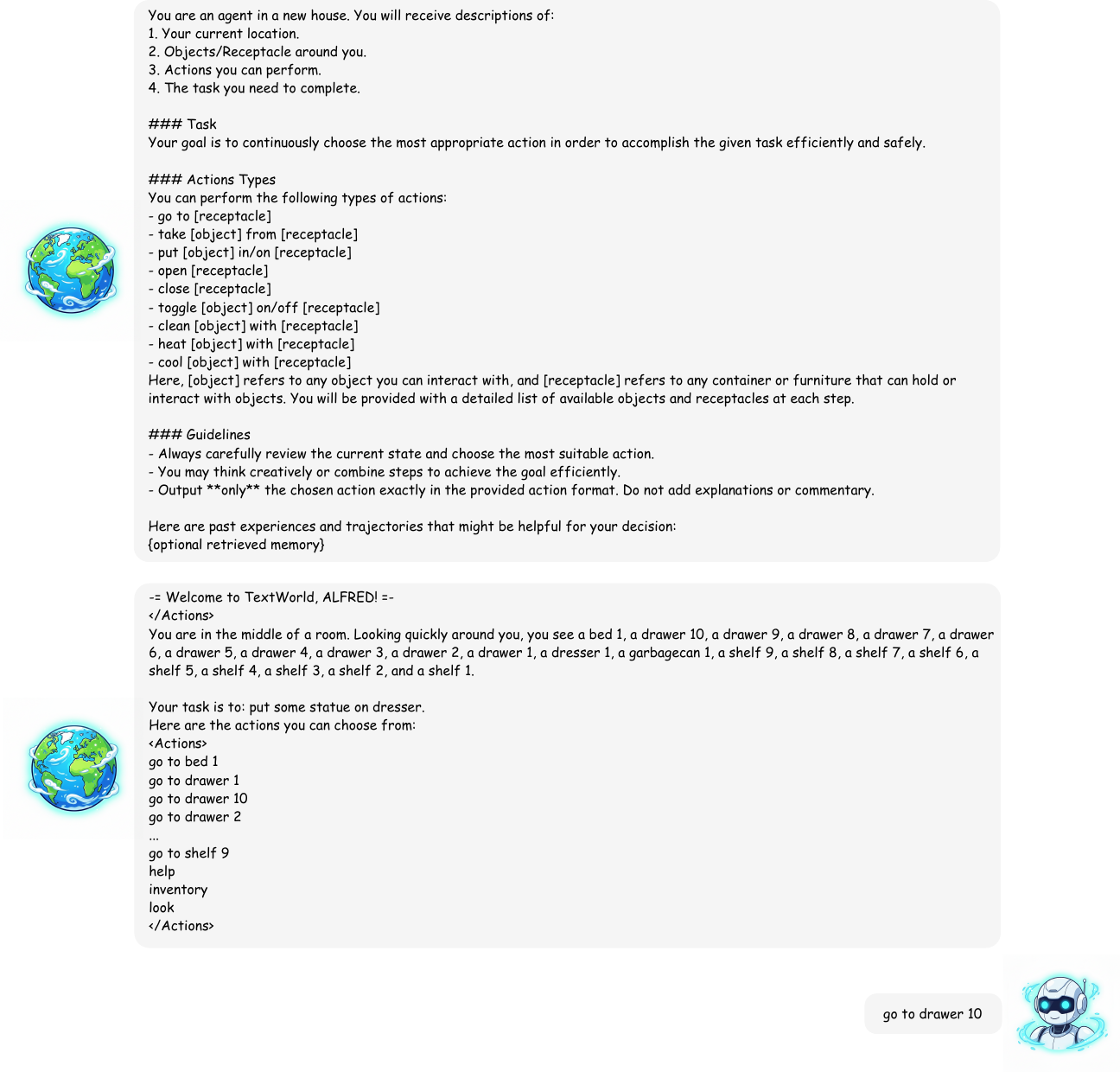}
  \caption{
  The example of system prompt, user prompt, and possible response given by the agentic system on ALFWorld.}
  \label{fig:alf_prompt}
\end{figure}

\noindent\textbf{ALFWorld.}\hspace{0.5em} ALFWorld is a text-based embodied environment derived from ALFRED \citep{Shridhar_2020_CVPR}, where agents interact with household environments through natural language instructions \citep{shridhar2021Alfworld}. 
Tasks in ALFWorld require agents to perform multi-step manipulations, such as picking, placing, opening, and toggling objects, often with dependencies between sub-tasks. 
We follow the standard ALFWorld benchmark configurations, as shown in \url{https://github.com/alfworld/alfworld/blob/master/configs/base_config.yaml}. It contains a \textit{train} dataset with 3553 tasks, a \textit{valid\_seen} with 140 tasks, and a \textit{valid\_unseen} dataset with 134 tasks. The \textit{valid\_seen} dataset only contains objects and rules that appear in the \textit{train} dataset, while the \textit{valid\_unseen} dataset contrains objects and rules that do not appear in \textit{train} dataset.
Each episode terminates either when the task is successfully completed or when a maximum step limit is reached, providing a binary reward of 1 for success and 0 for failure.
 In our experiment, we set a maximum step limit of 100. An example interaction is shown in \cref{fig:alf_prompt}.

\begin{figure}[htbp]
  \centering
  \includegraphics[width=1\textwidth]{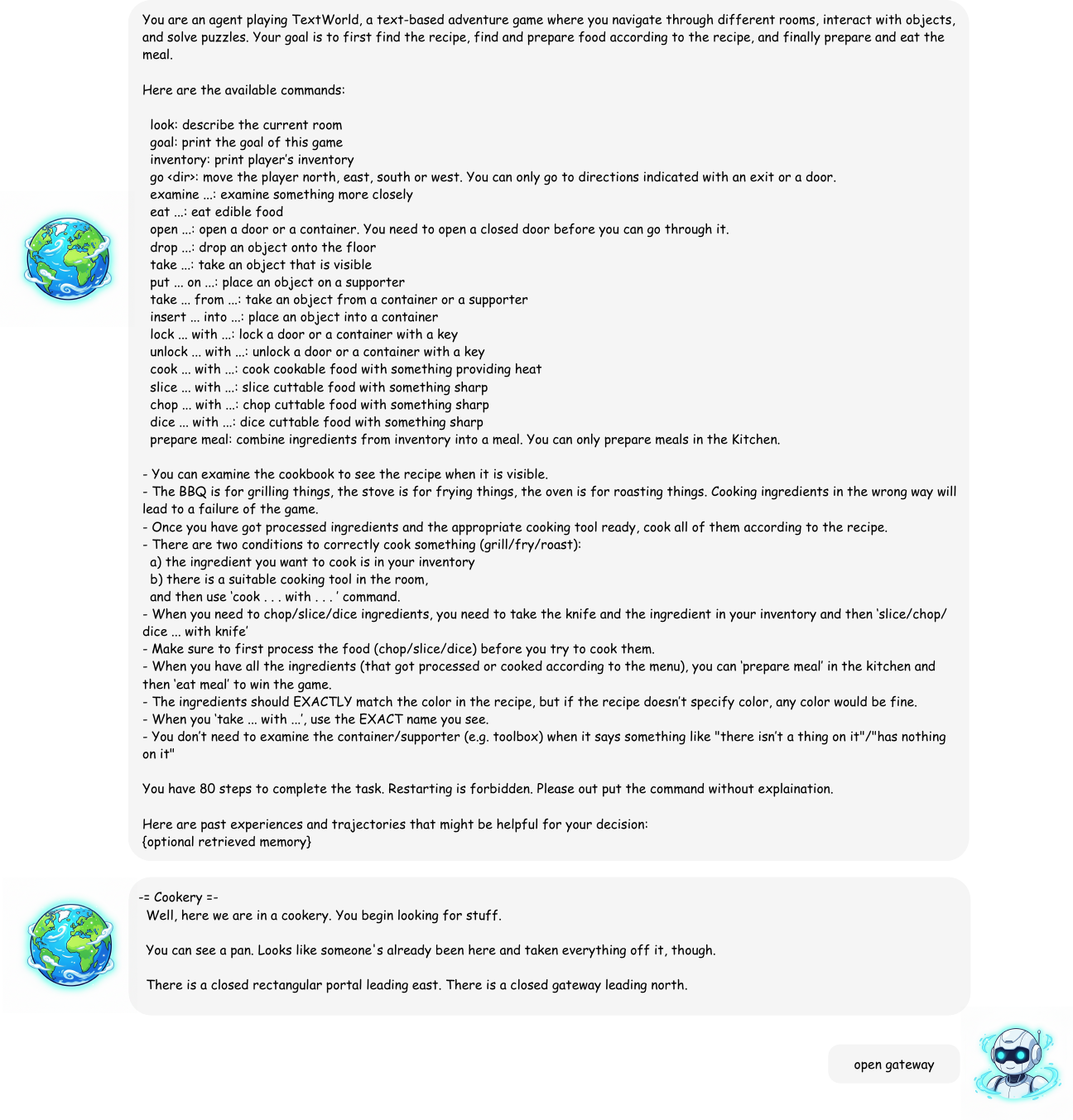}
  \caption{
  The example of system prompt, user prompt, and possible response given by the agentic system on \emph{Cooking} game in TextWorld.}
  \label{fig:textworld_prompt}
\end{figure}

\noindent\textbf{TextWorld.}\hspace{0.5em} 
TextWorld is a text-based interactive environment designed to evaluate sequential decision-making agents through natural language interaction \citep{textworld}. 
In TextWorld, the agentic system navigates and manipulates a partially observable world using textual commands, such as moving between rooms, examining objects, and performing object-centric actions.
We evaluate our method on the \emph{Treasure Hunter} and \emph{Cooking} games, using a total of 52 tasks across both games.
Each task requires the agentic system to complete a goal specified in natural language, often involving exploration, information gathering, and multi-step reasoning.
We perform evaluation using the BALROG benchmark \citep{balrog} and follow its standard configuration and prompting scheme from \citet{Lu2025International} (\url{https://github.com/balrog-ai/BALROG/blob/main/balrog/config/config.yaml}).
An episode terminates when the goal condition is satisfied or when the maximum step limit is reached, with a score ranging from 0 to 1 that measures the degree of task completion.
In our experiments, we set the maximum number of interaction steps to 80.
An example interaction is shown in \cref{fig:textworld_prompt}.

\begin{figure}[htbp]
  \centering
  \includegraphics[width=1\textwidth]{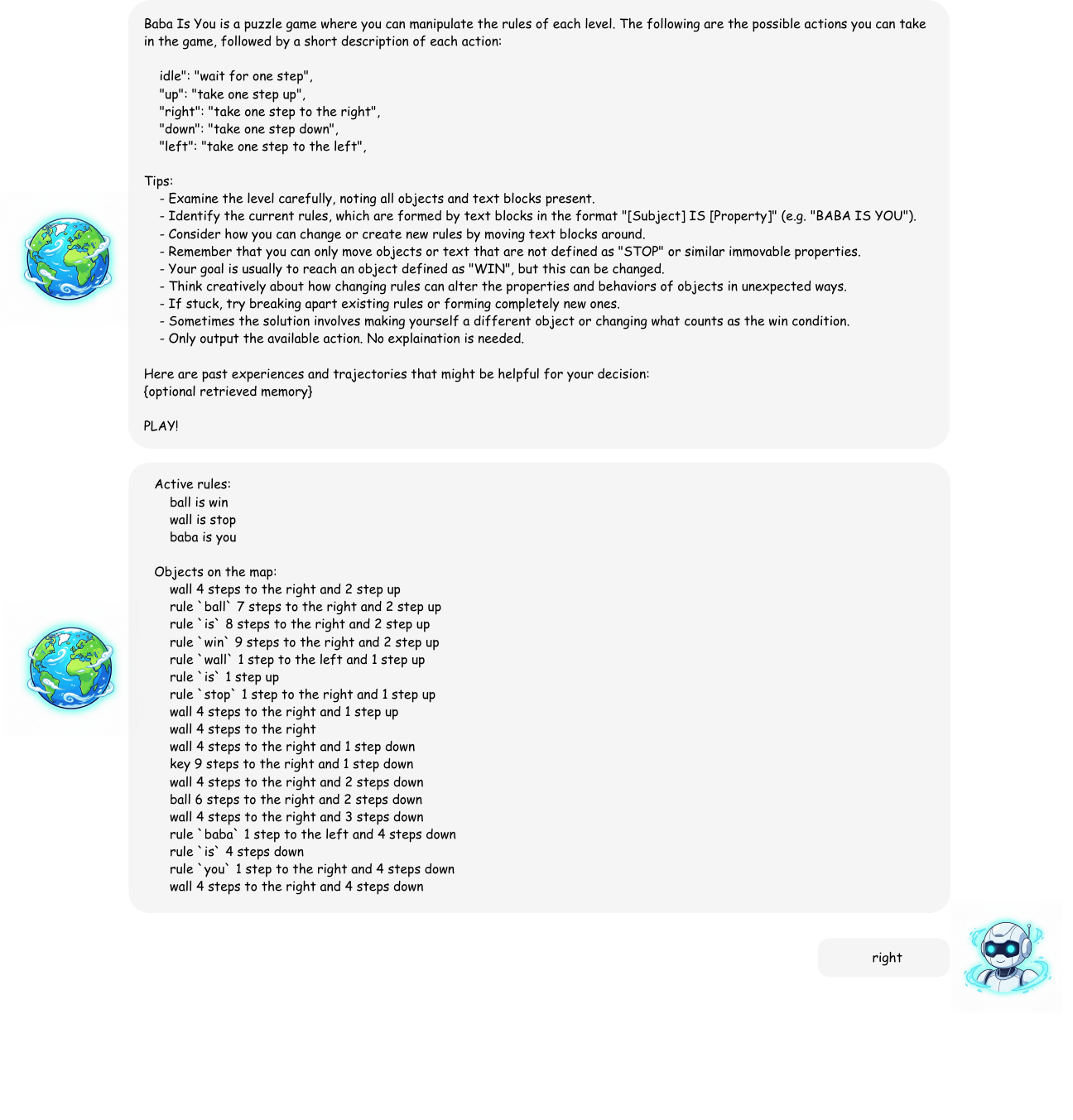}
  \caption{
  The example of system prompt, user prompt, and possible response given by the agentic system in Baba Is AI.}
  \label{fig:babaisai_prompt}
\end{figure}

\noindent\textbf{Baba Is AI.}\hspace{0.5em} 
Baba Is AI is a grid-based puzzle environment designed to evaluate sequential decision-making agents in worlds governed by explicitly manipulable symbolic rules \citep{cloos2024baba}. 
In Baba Is AI, an agentic system operates in a discrete grid world and interacts with objects by moving and pushing word tiles that define the rules of the environment.
The transition rules and goal conditions in Baba Is AI are specified as compositional natural language-like statements (e.g., \texttt{BABA IS YOU}, \texttt{FLAG IS WIN}) that can be modified through interaction.
Each task requires the agentic system to explore the grid, reason about the current rule set, and strategically alter rules to achieve the goal, often involving long-horizon planning and non-local dependencies.
We evaluate on Baba Is AI using BALROG \citep{balrog}, following its standard Baba Is AI benchmark configuration and prompting scheme (\url{https://github.com/balrog-ai/BALROG/blob/main/balrog/config/config.yaml}), using a total of 52 tasks.
An episode terminates when the win condition is satisfied under the current rule configuration or when the maximum step limit is reached, yielding a binary reward of 1 for success and 0 otherwise.
In our experiments, we set the maximum number of interaction steps to 20.
An example interaction is shown in \cref{fig:babaisai_prompt}.

\begin{figure}[htbp]
  \centering
  \includegraphics[width=1\textwidth]{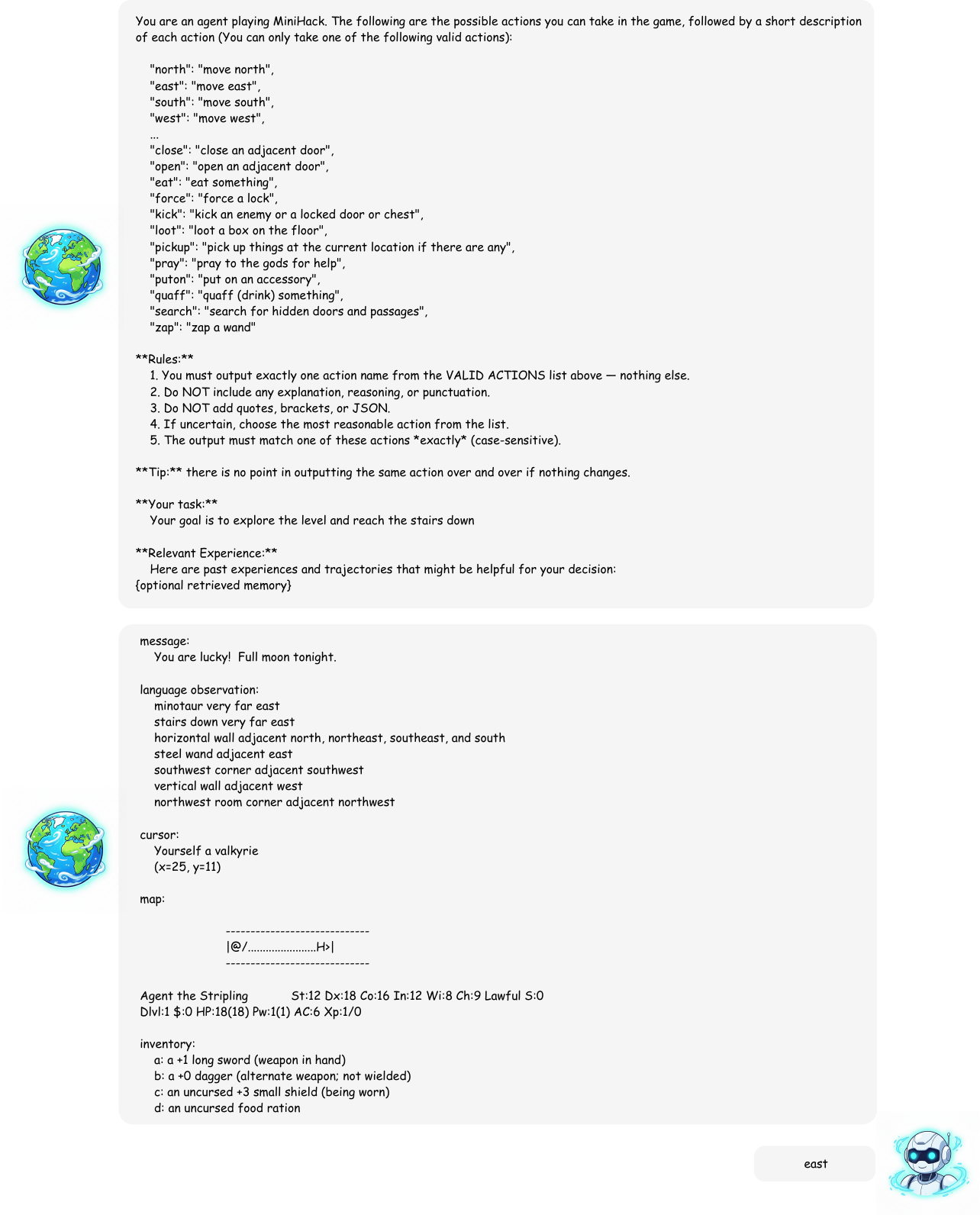}
  \caption{
  The example of system prompt, user prompt, and possible response given by the agentic system in MiniHack.}
  \label{fig:minihack_prompt}
\end{figure}

\noindent\textbf{MiniHack.}\hspace{0.5em} 
MiniHack is a grid-based sandbox environment built on top of the NetHack Learning Environment (NLE) \citep{samvelyan2021minihack, kuttler2020nle}, designed to evaluate sequential decision-making agents in procedurally generated dungeon worlds. 
We utilize \emph{Navigation} games, which require exploration, planning, and strategic resource management.
We evaluate on MiniHack using BALROG \citep{balrog}, following its standard MiniHack benchmark configuration and prompting scheme (\url{https://github.com/balrog-ai/BALROG/blob/main/balrog/config/config.yaml}), using a total of 114 tasks.
An episode terminates when the goal condition is satisfied or when the maximum step limit is reached, yielding a binary reward of 1 for success and 0 otherwise.
In our experiments, we set the maximum number of interaction steps to 100.
An example interaction is shown in \cref{fig:minihack_prompt}.

\newpage
\subsection{Baseline Details}
\label{app:baseline}
\noindent\textbf{Trajectory Retrieval.}\hspace{0.5em}
Trajectory Retrieval is a straightforward but strong baseline \citep{Park2023Generative, Xu2024International}.
Given a task description, we encode it into a vector representation and retrieve
the most similar past trajectory from the trajectory database based on cosine similarity.
Due to context length constraints, only the Top-1 retrieved raw trajectory is provided as memory to the agentic system.

\noindent\textbf{Dynamic Cheatsheet.}\hspace{0.5em}
Dynamic Cheatsheet (DC) maintains an external memory that tracks the successes and
failures of the agentic system \citep{suzgun2025DynamicCheatsheet}.
We adopt the cumulative Dynamic Cheatsheet (DC-Cu), which maintains a single global cheatsheet
throughout the evaluation of a memory design. We follow the prompts and workflows in \url{https://github.com/suzgunmirac/dynamic-cheatsheet}. 
During the Memory Collection Phase, trajectories are updated sequentially to
incrementally update the global cheatsheet.
In the Deployment Phase, the fixed global cheatsheet is provided to the agentic system to
evaluate the effectiveness of a shared global memory across tasks.

\noindent\textbf{ReasoningBank.}\hspace{0.5em} 
ReasoningBank maintains a structured memory schema in which all memory items are stored \citep{ouyang2025reasoningbank}.
Each memory item consists of three components:
(1) a title, which serves as an identifier summarizing the core strategy;
(2) a description, providing a one-sentence summary of the memory item; and
(3) the content, which records concise reasoning steps and insights.
During the Memory Collection Phase, memory items are constructed for each trajectory
by extracting either pitfalls or successful experiences, depending on the outcome
of the trajectory.
We follow the workflows and prompting strategies described in \citet{ouyang2025reasoningbank},
while using the ground-truth feedback of each trajectory.
During the Deployment Phase, we first retrieve the Top-1 trajectory corresponding
to the most similar task description from ReasoningBank, then use memory items
associated with the trajectory. The retrieved memory items are provided as memory for the agentic system.

\noindent\textbf{G-Memory.}\hspace{0.5em}
G-Memory uses a graph-based memory design, managing memory across three tiers: 
the insight, query, and interaction graphs \citep{zhang2025g}. 
We follow the workflows, prompts, and default settings provided in 
\url{https://github.com/bingreeky/GMemory}.
During the Memory Collection Phase, trajectories with task descriptions are first
provided to the query graph. They then traverse upward to the insight graph to
extract high-level insights, and downward to the interaction graph to identify
core interactions within the trajectories, forming the three-tier graph memory.
During the Deployment Phase, a task description is given to the query graph to
retrieve the Top-1 most similar trajectory. The retrieved trajectory is then
traversed upward and downward to identify the top-3 most similar insights and
core interactions. The retrieved insights and core interactions are provided as memory for the agentic system.

\subsection{Additional Details of Learning Process}
\label{app:learning_process}
\ouralgo is initialized with a memory design generated from scratch by the Meta Agent powered by \texttt{GPT-5/medium}. The generation of memory design follows Python Abstract Classes (Appendix~\ref{search_space}) and serves 
as the starting point for subsequent exploration. During generation, the Meta Agent 
is provided with \texttt{Provided Tools} containing minimal usage examples for 
invoking different agent roles (e.g., reasoning or conversational models), as well as 
embedding functions and database operations.

Starting from this initial memory design, \ouralgo performs 11 learning steps. 
At each step, memory designs are sampled from the archive, and the Meta Agent 
proposes, implements, and evaluates new memory designs. Each evaluated new design, together 
with its performance metrics and interaction logs, is then added to the memory design archive. 
This process repeats with the updated archive until the maximum number of learning 
iterations is reached.

\noindent\textbf{Sampling.}\hspace{0.5em} 
We perform the sampling process (Appendix~\ref{selection_process}) with 
$\alpha=0.5$ to balance the visit-time penalty, and $T=0.5$ to trade off between 
distribution smoothness and the probability of selecting higher-performing memory designs. 
Sampling is performed without replacement to maintain diversity in the selected subset 
while still favoring higher-probability designs, ensuring effective exploration and 
evaluation efficiency. As the memory design archive grows to more memory designs 
(e.g., 30 to 40), and only a small number of designs (five) are sampled at each 
step, the difference between sampling with and without replacement is expected to be 
negligible. Moreover, the purpose of sampling is to encourage exploration of memory designs that achieve 
high success rates but have been visited infrequently, while preserving a non-zero 
probability of sampling any memory design in the archive (Appendix~\ref{selection_process}). This is achieved through 
a weighted sampling scheme that balances performance and visitation frequency. 
Under this objective, both sampling with and without replacement are suitable, as 
they preserve the intended probability structure and maintain effective exploration 
behavior.
At each step, we sample up to five memory designs and perform parallel exploration of new memory designs.

\noindent\textbf{Greedy Search.}\hspace{0.5em} 
As an alternative to sampling in the learning process, we perform greedy search, 
always selecting the best memory design from the memory design archive at each learning step 
instead of sampling across all designs. This process continues until the total number 
of proposed designs matches that of the open-ended exploration. In our case, we perform 43 steps.
Finally, the best memory design in the memory design archive is selected as the learned design produced by greedy search. The results of greedy search are shown in \Cref{app:greedy_explore}.

\noindent\textbf{Ideate \& Planning.}\hspace{0.5em} 
The ideate \& planning procedure utilizes sampled memory designs along with their interaction logs 
to propose ideas and plans for new memory designs. The system prompt for the procedure 
remains fixed across all benchmarks and throughout the learning process 
(Appendix~\ref{prompts_meta}). For each sampled memory design, the relevant input information includes: 
(1) the source code of the current sampled memory design; 
(2) stratified-sampled interaction logs produced during the evaluation of the current sampled design; 
and (3) the overall success rate of the current sampled memory design.
Additionally, we include the following components to adapt 
to different benchmarks and situations for planning: 
(1) a one-shot example of plan proposed for a previous memory design, together 
with the associated performance of the previous design; 
(2) a benchmark description, depending on the benchmark being learned. The full benchmark description prompt is provided in \texttt{Benchmark Description}.

\begin{topblueprompt}{Benchmark Description}
\begin{lstlisting}[
  style=clean]
TASK_DESCRIPTION = {
        'alfworld': """The evaluation of downstream agent system is based on TextWorld: 
- The agent explores within a given space (rooms) and a predefined list of actions in order to accomplish a specified task goal. 
- Different data may involve **different rooms/spaces**, **different object placements**, or **entirely different task objectives**.
- The memory structure should strenthen following abilities of agent:
    - Spatial and State Reasoning: what are possible actions for each type of object?
    - Object-Centered Trajectory Learning: what did the agent do for previous same object management tasks or same location tasks?
""",
    'minihack':"""The evaluation of downstream agent system is based on MiniHack:
    - MiniHack is a collection of small game-like tasks where an agent sees a grid map of the environment, text and color representations, its own stats like health and position, and its inventory.  
    - Different games will have **different goals, maps and observations**. So goals and initial observations can change drastically across episodes, requiring **transferable knowledge**.
    - The memory structure should strenthen following abilities of agent:
        - Spatial and State Reasoning: Infer the environment from partial observations, remember explored areas, and navigate the map effectively.
        - Long-Horizon Goal Planning: Form multi-step strategies and consistently act toward the task objective despite sparse or delayed rewards.
        - Environmental Interaction and Risk Management: Correctly use actions and make safe, context-aware decisions around monsters, traps, and terrain.
    """,
    'textworld':"""The evaluation of downstream agent system is based on TextWorld:
    - The agent explores within a given space (rooms) and actions in order to accomplish a specified task goal. The agent may need to explore the space and get more progession. 
    - Done the task not neccessarily means the agent win the game. Whether wining or not depends on final reward(a progression percentage).
    - Different data may involve **different rooms/spaces**, **different object placements**, or **entirely different task objectives**.
    """,
    'babaisai':"""The evaluation of downstream agent system is based on BABAisAI:
    - In this gridworld game, agent interact with various objects and textual rule blocks to achieve specific goals.
    - Different data may involve **different rooms/spaces**, **different object placements**.
    - The agent goal is usually to reach an object defined as "WIN", but this can be changed by changing the rules.
    - The rules of the game can be manipulated and rearranged by the agent, creating a dynamic environment where agents must identify relevant objects and rules and then manipulate them to change or create new rules to succeed.
    - The memory structure should strenthen following abilities of agent:
        - ability to perceive all objects and text blocks in the environment, understand which rules are currently active or blocked, and evaluate how the environment state affects potential winning conditions.
        - ability to analyze and determine whether it is possible to win by modifying rule text blocks, and identify which rule combinations could achieve the goal.
        - ability to plan multi-step actions, including moving the character and pushing text blocks, to gradually construct or alter rules that lead to victory.
        - ability to flexibly adapt its strategy in response to dynamic rule and environment changes, handling potential obstacles or conflicts, and learn when modifying rules is more effective than simple movement.
        - ability to possess exploration and strategy transfer capabilities, discovering new ways to manipulate rules in unseen levels and applying prior experience to recognize when rule changes are advantageous.
    """
    }
\end{lstlisting}
\end{topblueprompt}

The output reflection and plans for the new memory design follows below schema:
\begin{topblueprompt}{Planning Schema}
\begin{lstlisting}[
  style=clean]
MEMO_ANALYSIS_OUTPUT_FORMAT = {
        "learned_from_suggestion_example": {
        "type": "string",
        "description": "Findings derived from the provided suggestion_example and improve_score. Bullet list of concrete factors (patterns) that made the suggestion succeed or fail. And principles to adopt when making future suggestions."
    },
    "trajectory_score_assessment": {
        "type": "array",
        "description": "analysis each retrieved module information based on current trajectories sampled and the benchmark scores.",
        "items": {
            "type": "object",
            "properties": {
                "label": {
                    "type": "string",
                    "enum": ["Useful", "Potentially Useful", "Irrelevant", "Empty/BadFormat"],
                    "description": "Categorization of the memory item's relevance based on whether the retrieved content actually helps the agent."
                },
                "how_it_can_help": {
                    "type": "string",
                    "description": "If Useful/Potentially Useful: short note how it could guide actions (subgoal, trap, object use...). If Irrelevant/Empty: reason (e.g., wrong keying, over-specific, missing summary, formatting)."
                }
            },
            "required": ["label", "how_it_can_help"]
        }
    },
    "content_quality_issues": {
        "type": "string",
        "description": "Detected content-level problems (duplicates, empty entries, serialization issues...). Why those harms retrieval or downstream planning.",
    },
    "structure_and_coherence": {
        "type": "string",
        "description": "Analysis of layer interactions, keying, and task-awareness. Which parts generalize, which are overfitted.",
    },
    "suggested_changes": {
        "type": "array",
        "description": "Based on all your analysis above, provide concrete change that can be applied on provided current memory structure code.",
        "items": {
            "type": "object",
            "properties": {
                "priority": {"type": "string", "enum": ["High", "Medium", "Low"], "description": "How urgent/impactful this is."},
                "what": {"type": "string", "description": "Precise description of what to change (code/pipeline/config)."},
                "why": {"type": "string", "description": "Link to observations/principles: why this addresses the problem."}
            },
            "required": ["priority", "what", "why"]
        }
    }
}
\end{lstlisting}
\end{topblueprompt}

\noindent\textbf{Implementation.}\hspace{0.5em} 
The Meta Agent implements new memory design using the proposed plan. Specifically, we provide several components to the Meta Agent for implementation: (1) \texttt{trajectory\_score\_assessment} and \texttt{suggested\_changes} in proposed plan, serve as instructions for implementation; (2) a one-shot trajectory example depending on the benchmarks as a reference format of possible input of the memory design; (3) Python Abstract Classes, which serve as the reference template for memory design implementation (Appendix~\ref{search_space}); and (4) provided tools for efficient implementation of new memory designs, as shown in \texttt{Provided Tools}.
\begin{topblueprompt}{Provided Tools}
\begin{lstlisting}[
  style=clean]
CHROMA_CHEETSHEET = """## Initialize Chroma DB

Use db = Chroma(embedding_function=Embedding()) to create the database. DO NOT use persist_dir.

### Add Memory: Adds new text entries to the database and returns their unique IDs. 

db.add_texts(
    texts: List[str],
    metadatas: Optional[Union[str, int, float, bool, None]] = None,
    ids: Optional[List[str]] = None
) -> List[str]

- metadatas must be **flat list**: each value must be a single primitive type (str, int, float, bool, or None).
- You cannot pass lists, nested dicts, or other complex objects.
- If you need to store structured data, serialize it to a JSON string:

### Retrieve Memory

db.similarity_search(
    query: str,
    k: int = 4
) -> List[Document]

return List[Document]: [
  Document(
    page_content="the agent found a key",
    metadata={"type": "item"}
  )
]

### Get by ID
db.get(
    ids: Optional[List[str]] = None
) -> Dict[str, List]

### Delete Memory
db.delete(
    ids: Optional[List[str]] = None
) -> None
    """
NXGRAPH_CHEETSHEET = """
NETWORKX GRAPH CHEATSHEET

Context: 
    import networkx as nx
    G = nx.Graph()

1. NODE OPERATIONS
- G.add_node(node, **attrs): Add a single node with optional attributes.
- G.add_nodes_from([n1, n2], **common_attrs): Add multiple nodes at once (shared attributes apply to all).
- G.remove_node(node): Remove a node and all edges connected to it.
- G.remove_nodes_from([n1, n2]): Remove multiple nodes.
- node in G: Check if a node exists.
- G.nodes: Get all nodes (NodeView).
- G.nodes[node]: Access node attributes as a dict.
- nx.set_node_attributes(G, {node: {"attr": value}}): Set attributes for nodes.

2. EDGE OPERATIONS
- G.add_edge(u, v, **attrs): Add an edge between two nodes.
- G.add_edges_from([(u, v), (x, y)], **attrs): Add multiple edges at once (shared attributes apply to all).
- G.remove_edge(u, v): Remove a single edge.
- G.remove_edges_from([(u, v), (x, y)]): Remove multiple edges.
- G.has_edge(u, v): Check if an edge exists.
- G.edges: Get all edges (EdgeView).
- G.edges[(u, v)]: Access edge attributes as a dict.
- nx.set_edge_attributes(G, {(u, v): {"weight": 1.0}}): Set attributes for edges.

3. TRAVERSAL / NEIGHBORHOOD
- G.neighbors(node): Get neighbors of a node.
- G.adj[node]: Get dict of neighbors with edge data.
- nx.shortest_path(G, source, target): Find one shortest path between nodes.
- nx.shortest_path_length(G, source, target): Get shortest path length.
- nx.all_simple_paths(G, source, target, cutoff): Generate all simple paths up to a cutoff length.
- nx.connected_components(G): Get connected components as node sets.
- G.subgraph([n1, n2, n3]): Extract a subgraph induced by given nodes.

4. ANALYSIS / CENTRALITY
- G.degree(node): Get degree (number of edges) for a single node.
- G.degree(): Get degree for all nodes (DegreeView).
- nx.degree_centrality(G): Compute degree centrality (dict of node -> score).
- nx.betweenness_centrality(G): Compute betweenness centrality.
- nx.pagerank(G): Compute PageRank scores for nodes.
- nx.clustering(G): Compute local clustering coefficient.
- nx.is_connected(G): Check if graph is connected.
- nx.number_connected_components(G): Count connected components.

5. UTILITIES
- G.copy(): Make a copy of the graph.
- G.clear(): Remove all nodes and edges.
- nx.to_dict_of_dicts(G): Convert graph to adjacency dict.
- nx.to_numpy_array(G): Get adjacency matrix as a NumPy array.
"""
TOOL_CHEETSHEET = """
TOOLS AVAILABLE:

1. Hire Agent
Class: Agent

- Purpose: Asynchronous wrapper around OpenAI Chat API, allows system/user prompts and optional JSON schema validation. Higher insights: this could be used to summarise information or gain new insights from them.
- Initialization: 
    from utils.hire_agent import Agent 
    agent = Agent(model: str, system_prompt: str, output_schema: Optional[Dict] = None)
- Key Methods:
    - agent.get_agent_config() -> Dict: Returns agent configuration and chat history.
    - await agent.ask(user_input: str, with_history: bool = False) -> Any:
        Sends a user message asynchronously and returns the model's response.
        If an output_schema is provided, response will be validated against it and returned as a JSON object.
- Usage Example:
    response = await agent.ask("Hello, how are you?")

-IMPORTANT: 
    - output_schema must be a **json schema**. Format example:
{
    "location": {
        "type": "string",
        "description": "The location to get the weather for"
    },
    "unit": {
        "type": ["string", "null"],
        "description": "The unit to return the temperature in",
        "enum": ["F", "C"]
    }
}

    - model could be "gpt-4.1", "gpt-4o-mini", based on whether you need reasoning ability(if reasoning ability is needed, gpt-4o-mini could be a better choice).

2. Embedding
Class: Embedding 

- Purpose: Async embedding manager for computing single or batch embeddings, with optional similarity calculation. Higher insights: this could be used to find connections between texts.
- Initialization:
    from utils.hire_agent import Embedding
    embedder = Embedding(model: str = "text-embedding-3-small", retries: int  = 3, retry_delay: float = 1.0)
- Key Methods:
    - await embedder.get_embedding(text: str) -> List[float]:
        Computes embedding for a single text string asynchronously.
    - await embedder.get_batch_embeddings(texts: List[str]) -> List[List[float]]:
        Computes embeddings for multiple texts asynchronously.
    - await Embedding.compute_similarity(emb1: List[float], emb2: List[float], metric: str = "cosine") -> float:
        Computes similarity between two embeddings asynchronously.
    - await Embedding.compute_one_to_group_similarity(emb: List[float], group_emb: List[List[float]], metric: str = "cosine") -> List[float]:
        Computes similarity between one embedding and a group of embeddings asynchronously.

- Notes:
    * All embedding calls automatically update a global token tracker.
    * Similarity functions support cosine similarity and run in parallel for efficiency.
"""
\end{lstlisting}
\end{topblueprompt}

As shown in \texttt{Provided Tools}, the Meta Agent is provided with the option of using 
\texttt{GPT-4o-mini} or \texttt{GPT-4.1} as FMs within the memory design, 
for example, when extracting insights from interaction trajectories. By explicitly including multiple model options (e.g., reasoning model or chat model) in the search space, 
the Meta Agent can adaptively select the most suitable model based on different reasoning 
and implementation requirements during memory design construction.
During testing, however, we consistently use \texttt{GPT-4o-mini} for all memory designs 
to ensure a fair comparison under matched model capacity.

We then perform a trial run to validate the implementation of the new memory design. 
The trial run samples 5 tasks from the learning set randomly, using 2 tasks for the 
Memory Collection Phase and 3 tasks for the Deployment Phase. If a runtime error occurs 
during the trial run, the Meta Agent engages in a reflection and debugging process, 
after which, the trial run is repeated on the debugged implementation. 
This loop of reflection, debugging, and trial runs iterates up to 3 times to ensure 
that the memory design implementation is free of bugs.

\noindent\textbf{Evaluation.}\hspace{0.5em} 
We integrate the new memory design with the agentic system powered by \texttt{GPT-5-nano/low}. We perform 
\texttt{general\_update} on each trajectory during the Memory Collection Phase. 
During the Deployment Phase, we use \texttt{general\_retrieve} to take the task 
description as input to retrieve relevant memory. 
The learning set is evenly split between the two phases. 
All tasks in the Deployment Phase are executed three times to compute the average 
success rate, reducing bias caused by variance in individual runs. For dataset splits, we use the first 30 tasks from the \textit{train} set in ALFWorld as the learning set. 
For TextWorld and Baba Is AI, the datasets are evenly split into learning and testing sets. 
For MiniHack, 30\% of the dataset is used as the learning set. 
The datasets for TextWorld, Baba Is AI, and MiniHack are sourced from BALROG \citep{balrog}.
We perform stratified sampling \citep{thompson2012sampling} to extract interaction logs after the Deployment Phase and store extracted logs in the memory design archive. 
When a memory design is sampled in future learning steps, the stratified logs allow the Meta Agent to reflect on both successes and failures 
without needing to examine all interaction logs.

\subsection{Additional Details of Testing}
\label{app:testing_details}
After learning memory designs on each benchmark, we perform testing under both static and dynamic modes for ALFWorld. For other benchmarks, we perform testing under static mode. For TextWorld and Baba Is AI, we use the second halves of the datasets for testing, respectively. For MiniHack, we use the remaining 70\% of the dataset for testing. All testing sets are split evenly for the Memory Collection and the Deployment Phase.

\noindent\textbf{Static Mode.}\hspace{0.5em} 
We first evaluate the best-learned memory designs and manual baselines on each benchmark using the agentic system powered by \texttt{GPT-5-nano/low}, the same model used during the learning process, to assess the effectiveness of the learned designs. 
We then test these memory designs with the agentic system powered by \texttt{GPT-5-mini/medium} to evaluate their transferability to a more capable agentic system. 
Additionally, for the ALFWorld \textit{valid\_seen} dataset, we test the learned best memory design alongside manual baselines while varying the number of trajectories used in the Memory Collection Phase from 0 to 70, using \texttt{GPT-5-mini/medium} as the foundational model in the agentic system.

\noindent\textbf{Dynamic Mode.}\hspace{0.5em} 
We evaluate the learned best memory design alongside manual baselines on ALFWorld using the agentic system powered by \texttt{GPT-5-mini/medium}. 
The first 70 tasks from the \textit{valid\_seen} dataset are used to generate trajectories for the Memory Collection Phase, 
while tasks from the \textit{valid\_unseen} dataset are used in the Deployment Phase, creating a gap between collected memory and tasks to be solved. 
During the Deployment Phase, the agentic system uses \texttt{general\_retrieve()} to access relevant memory for each new task. 
After completing a task, the interaction log is incorporated via \texttt{general\_update()}, producing a dynamic memory that evolves as the agentic system sequentially solves tasks.

\section{Result Details}
\subsection{Learning Processes on Other Benchmarks}
\label{app:meta_learn_results}
\begin{figure}[htbp]
  \centering
  \includegraphics[width=1\textwidth]{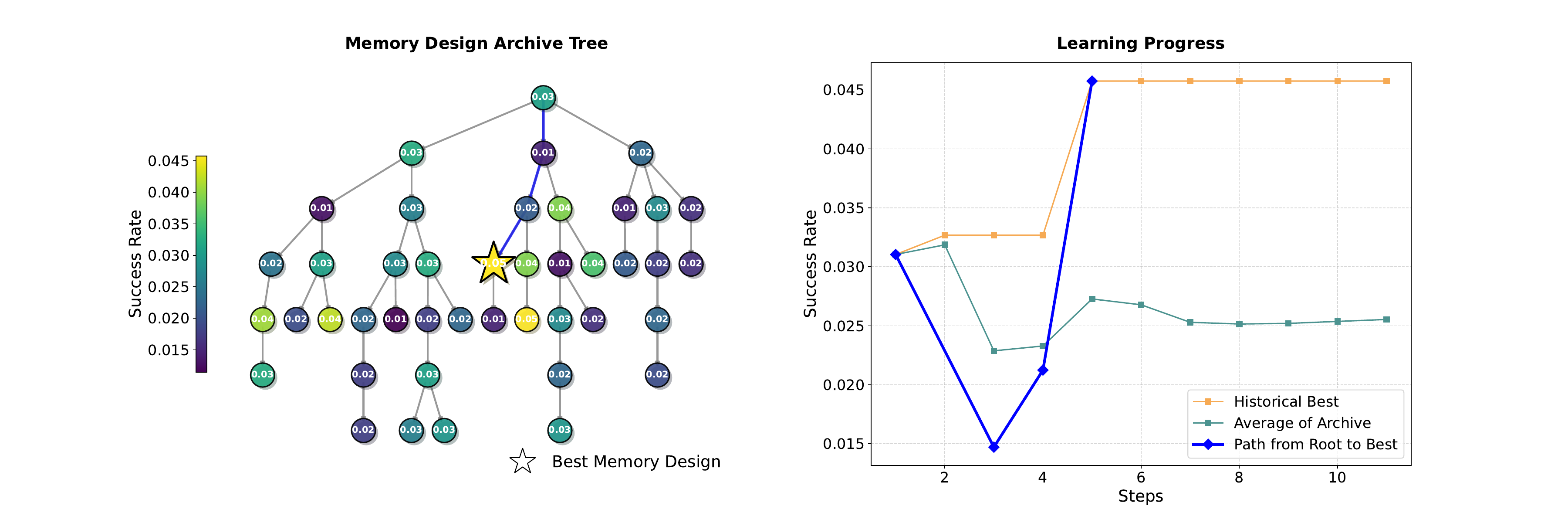}
  \captionsetup{aboveskip=5pt, belowskip=5pt}
  \caption{The learning process of \ouralgo on TextWorld, using \texttt{GPT-5-nano} as the FM in an agentic system.
    \textbf{Left:} The memory design archive tree, where each node represents a memory design. Node colors indicate the success rate, and edges indicate that each child node is derived from its parent. The memory design with the highest success rate is used as the final learned memory design.
    \textbf{Right:} The step-wise learning progress. \ouralgo progressively discovers memory designs by building on an ever-growing archive of previous discoveries. The path from the root memory design to the best memory design highlights the importance of open-ended exploration, where designs with moderate success rates serve as stepping stones toward optimal solutions.}
  \label{fig:textworld_reward_tree}
\end{figure}
\begin{figure}[htbp]
  \centering
  \includegraphics[width=1\textwidth]{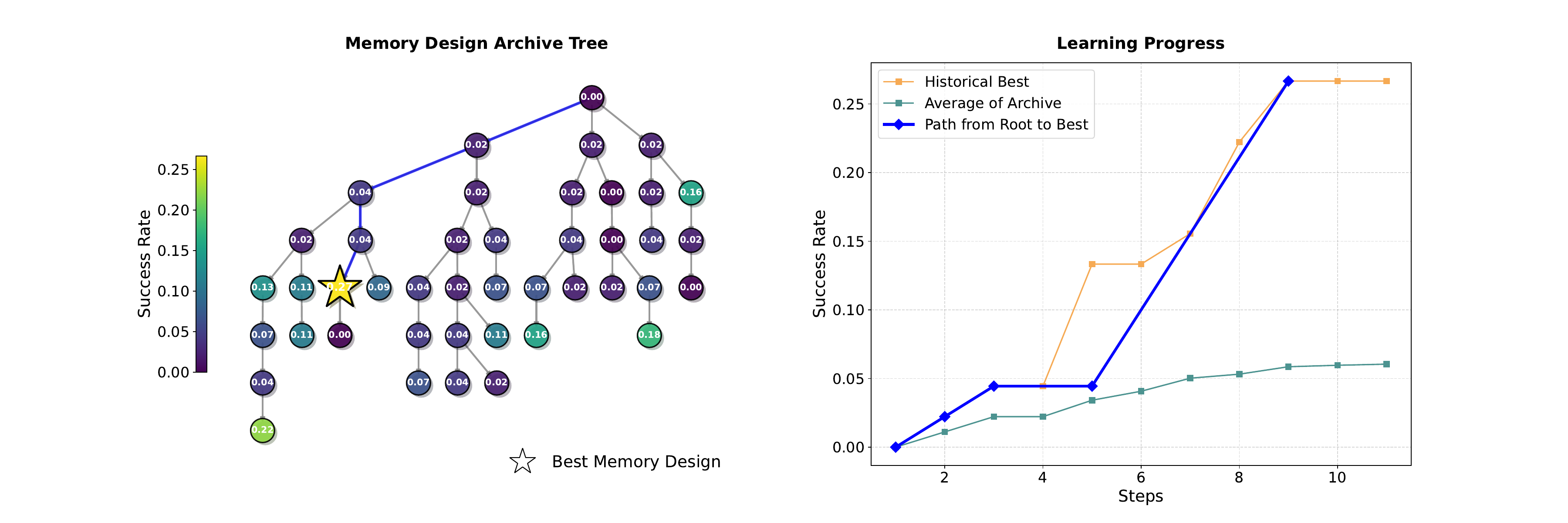}
  \captionsetup{aboveskip=5pt, belowskip=5pt}
  \caption{
      The learning process of \ouralgo on ALFWorld, using \texttt{GPT-5-nano} as the FM in an agentic system.
    \textbf{Left:} The memory design archive tree, where each node represents a memory design. Node colors indicate success rate, and edges indicate that each child node is derived from its parent. The memory design with the highest success rate is used as the final learned memory design.
    \textbf{Right:} The step-wise learning progress. \ouralgo progressively discovers memory designs by building on an ever-growing archive of previous discoveries. The path from the root memory design to the best memory design highlights the importance of open-ended exploration, where designs with moderate success rates serve as stepping stones toward optimal solutions.}
  \label{fig:alfworld_reward_tree}
\end{figure}
\begin{figure}[htbp]
  \centering
  \includegraphics[width=1\textwidth]{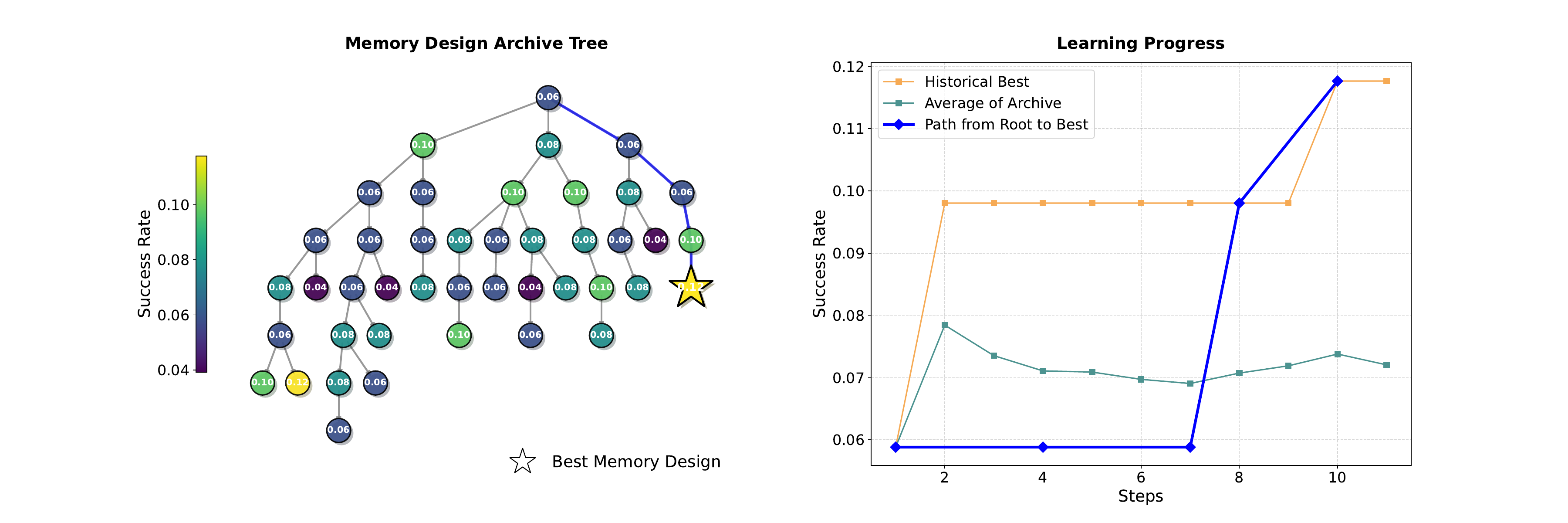}
  \captionsetup{aboveskip=5pt, belowskip=5pt}
  \caption{The learning process of \ouralgo on MiniHack, using \texttt{GPT-5-nano} as the FM in an agentic system.
    \textbf{Left:} The memory design archive tree, where each node represents a memory design. Node colors indicate success rate, and edges indicate that each child node is derived from its parent. The memory design with the highest success rate is used as the final learned memory design.
    \textbf{Right:} The step-wise learning progress. \ouralgo progressively discovers memory designs by building on an ever-growing archive of previous discoveries. The path from the root memory design to the best memory design highlights the importance of open-ended exploration, where designs with moderate success rates serve as stepping stones toward optimal solutions.}
  \label{fig:minihack_reward_tree}
\end{figure}
\newpage
\subsection{Results of Greedy Exploration}
\label{app:greedy_explore}
\begin{figure}[htbp]
  \centering
  \includegraphics[width=0.8\textwidth]{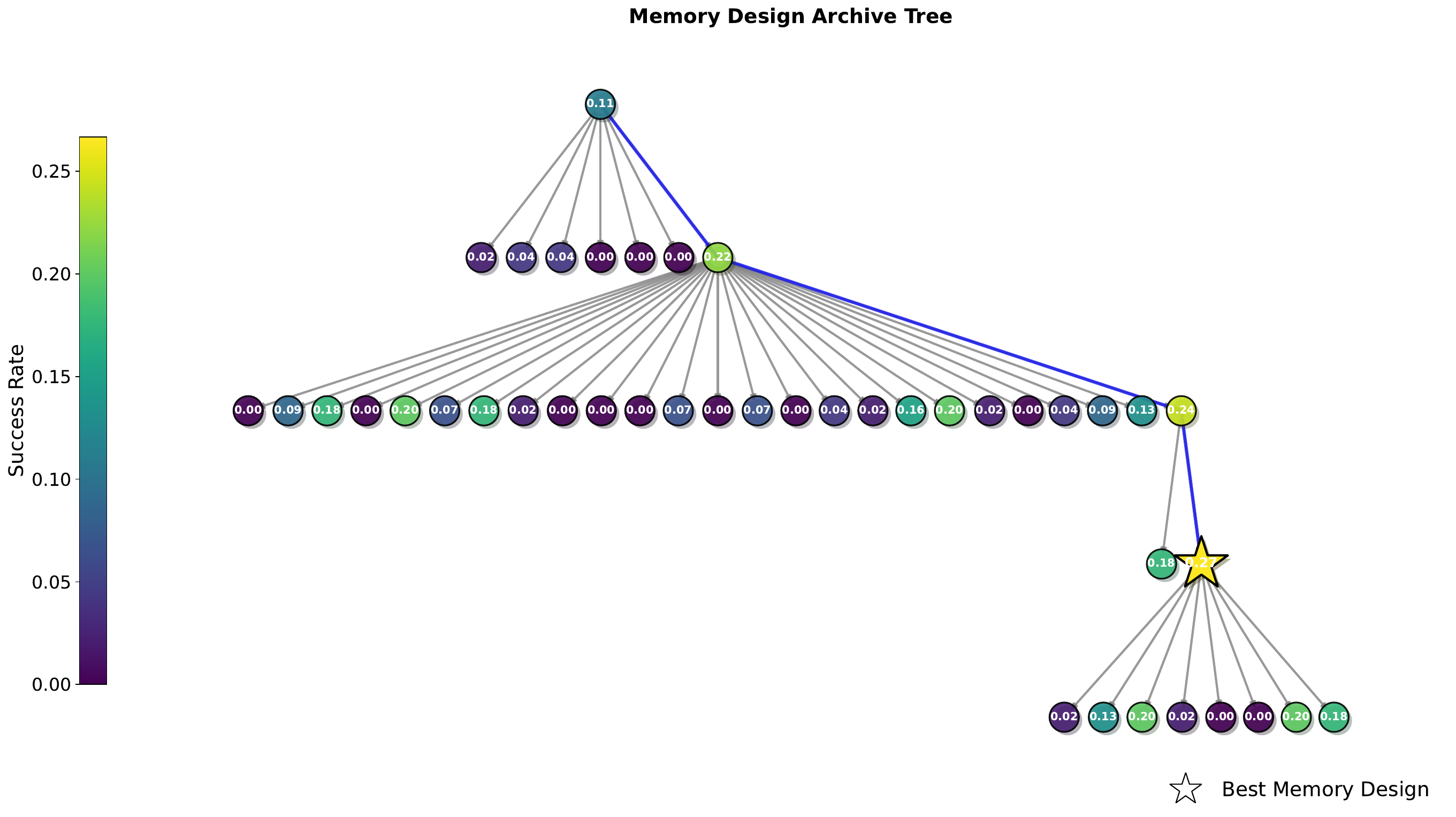}
  \caption{The memory design archive tree of the greedy learning process on ALFWorld, using \texttt{GPT-5-nano} as the FM in the agentic system. Each node represents a memory design. Node colors indicate success rate, and edges indicate that each child node is derived from its parent. The memory design with the highest success rate is used as the final learned memory design. In the greedy learning process, each new memory design is developed based on the previous design that achieved the highest success rate.}
  \label{greedy_reward_tree}
\end{figure}

We also conduct a greedy memory design sampling baseline on ALFWorld, where at each learning step we only select the historically best-performing memory design for further exploration (Appendix~\ref{app:learning_process}).
To ensure a fair comparison, this greedy strategy is evaluated under the same exploration budget, i.e., it explores the same total number of memory designs as our method.
The resulting exploration tree is shown in \cref{greedy_reward_tree}.

\ouralgo produces learned memory designs that achieve better performance and transferability (\cref{tab:greedy_vs_openended}) compared to Greedy Search, highlighting the importance of open-ended exploration patterns to explore better designs.

\begin{table}[htbp]
\centering
\captionsetup{aboveskip=5pt, belowskip=5pt}
\caption{Comparison of success rate (Mean $\pm$ SE) in percentage using Greedy Search and \ouralgo on ALFWorld. The standard error is calculated over three runs of the Deployment Phase. Testing is conducted with \texttt{GPT-5-nano} and \texttt{GPT-5-mini} as foundation models in the agentic system.}
\label{tab:greedy_vs_openended}
\footnotesize
\setlength{\tabcolsep}{6pt}
\begin{tabular}{l c c } 
\toprule
\textbf{Foundation Model} & \textbf{Greedy Search} & \textbf{\ouralgo (Ours)}\\
\midrule
GPT-5-nano & 11.9 $\pm$ 0.5 & \textbf{12.4 $\pm$ 0.5}\\
GPT-5-mini & 77.1 $\pm$ 0.8 & \textbf{87.1 $\pm$ 1.4}\\
\bottomrule
\end{tabular}
\vskip 10pt
\end{table}

\newpage
\subsection{Example Results of Learned Memory Designs}
\label{app:learned_design}

\lstinputlisting[
  style=frame,
  language=Python,
  caption={The Best-learned memory design in MiniHack.},
  label={lst:minihack},
  basicstyle=\ttfamily\footnotesize, 
  breaklines=true,
  breakatwhitespace=false, 
  xleftmargin=0.025\columnwidth, 
  xrightmargin=0.025\columnwidth, 
  linewidth=0.95\columnwidth 
]{code/minihack.py}

\begin{lstlisting}[
  style=frame,
  language=python,
  caption={Example of the retrieved knowledge context during the Deployment Phase in MiniHack.},
  label={lst:minihack_output}
    basicstyle=\ttfamily\footnotesize, % 1. 缩小字体，防止代码过宽
  breaklines=true,
  breakatwhitespace=false, % 2. 关键：允许在特殊符号处换行，而不仅仅是空格
  xleftmargin=0.025\columnwidth, % 3. 左边留 2.5% 的空隙
  xrightmargin=0.025\columnwidth, % 4. 右边留 2.5% 的空隙
  linewidth=0.95\columnwidth % 5. 总宽度 95%，配合左右 margin 实现居中
]
memo_retrieved = {
"task_overview": {
    "goal": "Your goal is to get as far as possible in the game.",
    "role": "rogue",
    "actions": [
    "apply",
    "east",
    "north",
    "northeast",
    "northwest",
    "pickup",
    "south",
    "southeast",
    "southwest",
    "west"
    ],
    "map_size": {
    "width": 80,
    "height": 21
    },
    "initial_position": {
    "x": 41,
    "y": 18
    },
    "initial_entities": [
    "wall"
    ],
    "inventory_items": [
    "short sword",
    "dagger",
    "leather armor",
    "potion of sickness",
    "lock pick",
    "empty sack"
    ],
    "local_topology": {
    "walls_at": [
        "east",
        "north",
        "northeast",
        "south",
        "southeast",
        "southwest"
    ],
    "doors_at": [],
    "stairs_up_at": [],
    "stairs_down_at": [],
    "boulder_at": [],
    "fountain_at": [],
    "other": "{}"
    },
    "map_neighbors": {
    "north": ".",
    "northeast": ".",
    "east": ".",
    "southeast": ".",
    "south": ".",
    "southwest": ".",
    "west": ".",
    "northwest": "."
    },
    "passable_dirs": [
    "north",
    "northeast",
    "east",
    "southeast",
    "south",
    "southwest",
    "west",
    "northwest"
    ],
    "initial_movement_suggestions": {
    "blocked_dirs": [],
    "hazard_dirs": [],
    "suggested_first_moves": [
        "east",
        "north",
        "northeast",
        "northwest"
    ],
    "nearby_objectives": {
        "topology": {}
    },
    "fallback_actions": []
    },
    "action_mask": {
    "allowed": [
        "east",
        "north",
        "northeast",
        "northwest",
        "south",
        "southeast",
        "southwest",
        "west"
    ],
    "blocked": [],
    "hazard": [],
    "fallback": []
    },
    "similar_cases": [
    {
        "case_id": "34b90c20-541c-4148-bca7-6df9979f2b3e",
        "reward": 1.0,
        "outcome": "failure_or_partial",
        "lesson": "Always assess the surrounding topology for passable routes before acting. Prioritize movement toward open paths to maximize progress."
    },
    {
        "case_id": "30541a20-6805-4222-883c-3ea8537a4bd8",
        "reward": 1.0,
        "outcome": "failure_or_partial",
        "lesson": "Always assess your surroundings for walls before moving. Avoid repetitive actions that lead to dead ends and seek alternative routes to progress."
    }
    ]
},
"suggested_strategies": {
    "plan_outline": {
    "plan_title": "Navigating the Grid to Progress",
    "ordered_steps": [
        "Assess initial position and adjacent walls.",
        "Determine the safest path avoiding walls.",
        "Utilize available inventory for movement and defense.",
        "Plan movement towards open spaces or unexplored areas.",
        "Execute movement step by step, keeping track of the path.",
        "Monitor health and inventory status.",
        "Adapt strategy based on encountered obstacles."
    ],
    "pitfalls_to_avoid": [
        "Rushing movements without assessing surroundings.",
        "Ignoring the potential benefits of inventory items.",
        "Failing to track health status leading to unplanned deaths.",
        "Becoming fixated on one direction despite walls blocking the way."
    ],
    "key_checks": [
        "Confirm no walls block planned movement direction.",
        "Reassess health and inventory after each turn.",
        "Ensure there are no dead ends or traps in the path ahead.",
        "Check if any enemies appear during movement."
    ]
    },
    "bulleted_tips": [
    "Survey adjacent passable tiles to uncover new paths.",
    "Avoid repeatedly moving in circles; change direction if blocked.",
    "Focus on exploring directions that lead to visible objectives.",
    "Utilize your lock pick to open potential paths hidden behind walls.",
    "Keep your inventory organized; prioritize items based on immediate needs."
    ]
},
"spatial_priors": [
    "wall -> blocks_movement (support=280.5)"
],
"risk_and_interactions": [
    "Explore passable directions; avoid blocked paths.",
    "Keep escape routes clear; don't get cornered.",
    "Use the potion of sickness when HP is low for risky moves."
],
"reflex_tips": [
    "Avoid bumping into blocked directions: east, north, northeast, south, southeast, southwest. Pivot to open tiles.",
    "Walls/bars block movement; navigate around instead of repeating bumps."
],
"retrieval_metadata": {
    "task_id": "minihack:643759935e128b64511fe4c27717958dc1ebf583",
    "knowledge_query": "Your goal is to get as far as possible in the game. | role:rogue | entities:wall | inv:short sword,dagger,leather armor,potion of sickness,lock pick,empty sack | topo:walls:east,north,northeast,south,southeast,southwest; up:; down:; boulder:; fountain: | neighbors:N:.; NE:.; E:.; SE:.; S:.; SW:.; W:.; NW:. | passable:north,northeast,east,southeast,south,southwest,west,northwest | hazards:"
}
}
\end{lstlisting}

\end{document}